\documentclass[letterpaper]{article} 
\usepackage{aaai2026}  
\usepackage{times}  
\usepackage{helvet}  
\usepackage{courier}  
\usepackage[hyphens]{url}  
\usepackage{graphicx} 
\usepackage{natbib}  
\usepackage{caption} 
\usepackage{adjustbox}
\usepackage{booktabs}
\usepackage{amsfonts}
\usepackage{multirow}
\usepackage{amsmath}
\usepackage{subcaption}
\usepackage{makecell}
\usepackage{algorithm}
\usepackage{algorithmic}

\usepackage{newfloat}
\usepackage{listings}
\DeclareCaptionStyle{ruled}{labelfont=normalfont,labelsep=colon,strut=off} 
\floatstyle{ruled}
\newfloat{listing}{tb}{lst}{}
\floatname{listing}{Listing}
\title{Resource Efficient Sleep Staging via Multi-Level Masking and Prompt Learning}
\author{
    Lejun Ai\textsuperscript{\rm 1,2},
    Yulong Li\textsuperscript{\rm 1},
    Haodong Yi\textsuperscript{\rm 1},
    Jixuan Xie\textsuperscript{\rm 1},
    Yue Wang\textsuperscript{\rm 1},
    Jia Liu\textsuperscript{\rm 3},\\
    Min Chen\textsuperscript{\rm 1,2}\thanks{Corresponding authors.},
    Rui Wang\textsuperscript{\rm 3,4}\footnotemark[1]
}
\affiliations{
    \textsuperscript{\rm 1}School of Computer Science and Engineering, South China University of Technology, Guangzhou, China,\\
    \{cs\_ailejun, csyulongli, cshaodongyi, csjixuanxie, csyuewang\}@mail.scut.edu.cn,\\
    \textsuperscript{\rm 2}Pazhou Laboratory, Guangzhou, China, minchen@ieee.org\\
    \textsuperscript{\rm 3}School of Computer Science and Technology, Huazhong University of Technology, Wuhan, China,\\
    \{jialiu0330, ruiwang2020\}@hust.edu.cn,\\
    \textsuperscript{\rm 4}Guangdong HUST Industrial Technology Research Institute, Dongguan, China

}

\usepackage{bibentry}

\begin{document}

\maketitle

\begin{abstract}
Automatic sleep staging plays a vital role in assessing sleep quality and diagnosing sleep disorders. Most existing methods rely heavily on long and continuous EEG recordings, which poses significant challenges for data acquisition in resource-constrained systems, such as wearable or home-based monitoring systems. In this paper, we propose the task of resource-efficient sleep staging, which aims to reduce the amount of signal collected per sleep epoch while maintaining reliable classification performance. To solve this task, we adopt the masking and prompt learning strategy and propose a novel framework called \textbf{Mask-Aware Sleep Staging (MASS)}. Specifically, we design a multi-level masking strategy to promote effective feature modeling under partial and irregular observations. To mitigate the loss of contextual information introduced by masking, we further propose a hierarchical prompt learning mechanism that aggregates unmasked data into a global prompt, serving as a semantic anchor for guiding both patch-level and epoch-level feature modeling. MASS is evaluated on four datasets, demonstrating state-of-the-art performance, especially when the amount of data is very limited. This result highlights its potential for efficient and scalable deployment in real-world low-resource sleep monitoring environments.
\end{abstract}

\begin{links}
    \link{Code}{https://github.com/AnsonAiTRAY/MASS}
    \link{Extended version}{https://arxiv.org/abs/2511.06785}
\end{links}

\begin{figure}[!ht]
    \centering
    \includegraphics[width=\linewidth]{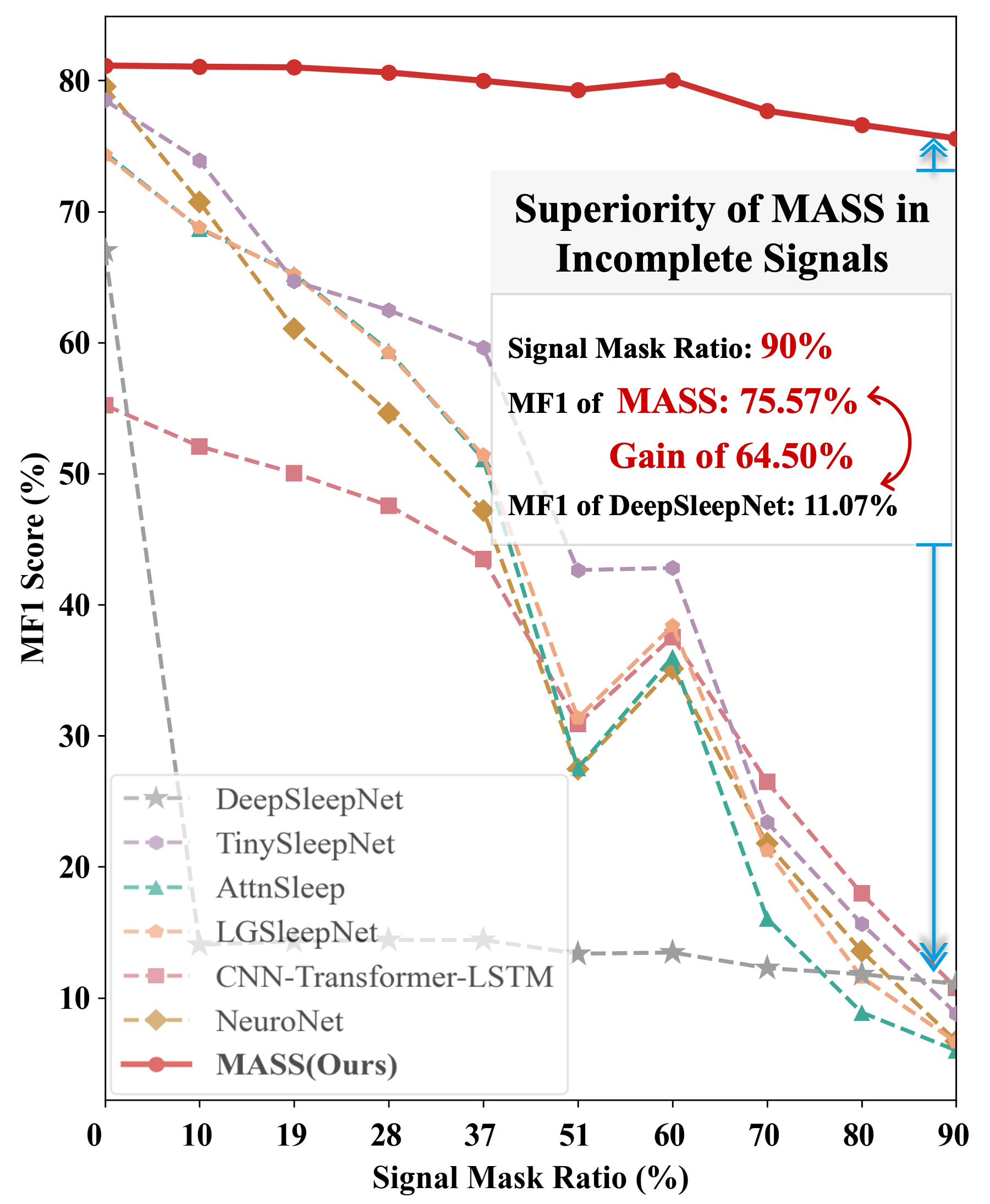}
    \caption{Comparison of macro-F1 scores in resource-efficient scenarios. When the signal mask ratio increases, performance of models rely on complete training data decrease significantly. Even though there is redundancy in the complete signal, they cannot directly utilize such a small amount of signal for inference. Our MASS framework based on multi-level masking and prompt learning, which can focus on features under a small amount of information and achieve reliable and resource-efficient sleep staging.}
    \label{fig:acc}
\end{figure}

\section{Introduction}
\label{introduction}
Sleep occupies one-third of a person's life and plays a vital role in both physical and mental health~\cite{Scott2021Improving}~\cite{Fernandez-Mendoza2013Insomnia}~\cite{cappuccio2010sleep}~\cite{irwin2019nature}~\cite{IRWIN2019296}. Sleep staging is a key method for the evaluation of sleep quality and the diagnosis of sleep disorders~\cite{Boostani2017A}~\cite{Memar2018A}. According to the standard criteria proposed by American Academy of Sleep Medicine (AASM), a 30-second electroencephalogram (EEG) signal, known as a sleep epoch~\cite{berry2017aasm}, can be classified into one of six sleep stages: Wake (W), rapid eye movement (REM), and four non-REM stages (N1, N2, N3, and N4)~\cite{berry2012aasm}. Traditionally, sleep staging has relied on manual examinations conducted by physicians, which is labor-intensive and constrained to specific clinical environments. Advances in flexible wearable sleep monitoring systems offer a potential approach for reducing dependence on human expertise and specific environmental settings~\cite{qiu2024reliable}~\cite{chen2022fabric}~\cite{chen2022multifunctional}. These systems mainly employ neural networks on edge servers for automatic sleep staging, with flexible fabric sensors and miniature amplifiers to enable easier and less intrusive signal collection. Together, these research make flexible wearable sleep monitoring systems particularly well-suited for home environments due to their lightweight design and automated functionality.\par
However, flexible wearable sleep monitoring systems face significant challenges of resource limitations~\cite{pan2020jmir}. In particular, the compact design and the deployment in home environments impose strict constraints on battery capacity. These constraints critically hinder the system's long-term sustainability for sleep monitoring. A major underlying cause of this limitation is the high data acquisition and pre-processing overhead required by existing systems. Traditional sleep staging methods typically depend on the availability of complete and continuous EEG recordings of each sleep epoch to achieve reliable performance. This strong reliance makes them particularly vulnerable in resource-constrained scenarios, where reducing the amount of input data often leads to a substantial drop in classification accuracy. To overcome this challenge, we propose a novel neural network model that maintains robust performance even when using only 10\% of the original signal, thereby significantly enhancing the efficiency and practicality of resource-limited wearable sleep monitoring systems.\par
In this paper, we introduce \textbf{Mask-Aware Sleep Staging (MASS)}, a novel and resource-efficient framework designed to reduce the dependency on full-length EEG signals for sleep staging. During training and inference, MASS applies both intra-epoch and inter-epoch random masking to the EEG signals. This multi-level masking strategy encourages robust representation learning by forcing the model to focus on learning local, partially observed signals rather than relying on the entire signal. MASS further incorporates a global prompt learning mechanism, which models the overall relationships among all visible epochs and their internal visible patches. The global prompt is integrated into both intra-epoch and inter-epoch encoding processes to guide the local feature extraction. As a result, MASS achieves state-of-the-art performance in sleep staging while substantially reducing signal requirements as shown in Fig.~\ref{fig:acc}. This design naturally supports on-device acquisition control, enabling sampling to pause during masked segments to reduce power without materially degrading staging accuracy. At the same time, this is also achieved without increasing computational complexity. To the best of our knowledge, MASS represents the first neural network-based approach that explicitly addresses the challenge of resource constraints in wearable sleep monitoring by optimizing for data efficiency.\par
The main contributions of the present work are as follows:
\begin{itemize}
    \item The problem of \textbf{resource-efficient sleep staging} in flexible wearable systems is formally defined, with a focus on reducing signal acquisition demands under strict hardware constraints, an issue rarely explored in prior neural network-based approaches.
    \item \textbf{Mask-Aware Sleep Staging (MASS)} is proposed, a novel framework that introduces multi-level masking strategy to exploit patch-level and epoch-level correlations, and global prompt learning mechanism to guide the multi-level feature learning. This hybrid architecture can enable accurate sleep staging with only \textbf{10\%} of the original EEG signal.
    \item We conducted experiments on four public datasets with different signal integrity, proving that our MASS method can achieve stable and reliable sleep staging even under high signal masking conditions, far surpassing other state-of-the-art methods.
\end{itemize}\par

The remainder of the paper is structured as follows: Section II reviews related work in sleep staging and wearable systems. Section III introduces the proposed Mask-Aware Sleep Staging method, including its architecture and resource-efficient learning strategy. Section IV describes the datasets and experiment results. Section V presents discussion and conclusion.\par

\section{Related Work}
In the field of sleep staging with EEG signals, the most widely used approach is convolutional neural network (CNN), transformer and multi-head attention-based models have also been integrated with CNN to some extent~\cite{vaswani2017attention}. Supratak et al.~\cite{7961240} used a dual-stream CNN network to learn EEG signal representation and employed bidirectional long short-term memory to learn the contextual dependencies between adjacent epoch features. Eldele et al.~\cite{9417097} applied a multi-resolution CNN with adaptive feature recalibration model to learn multi-frequency features, and implemented a multi-head attention mechanism to capture the temporal dependencies within the extracted features. Shen et al.~\cite{10195897} used a CNN kernel to extract local features and a transformer to extract global features. Shen et al.~\cite{shen2024robust} further introduces a contrastive imagination framework to perform sleep staging over incomplete multimodal signals by learning cross-modal representations and reconstructing missing information. Phyo et al.~\cite{9877908} proposed an attention-based module to capture salient waveforms and two auxiliary tasks to classify confusing stages accurately during transitioning epochs. In summary, CNN combined with transformer or LSTM is a typical approach for sleep staging based on single-channel EEG signals. \par
Sleep monitoring is indispensable to diagnosis and treatment of sleep disorders. Works about resource-efficient monitoring are classified into two aspects: algorithms in software and work modes in hardware. In terms of software, some proposed approaches such as TinySleepNet~\cite{supratak2020tinysleepnet} take computational resources into accounts and manage to reduce the number of model parameters. In addition, CNN-Transformer-LSTM~\cite{pham2023automatic} has fewer parameters, which is beneficial for resource-efficient computation. In terms of hardware, several commercially available EEG monitors, such as ADS1299~\cite{ADS1299} or ADS1294~\cite{ADS1294}, support microsecond-level switching between normal collection mode and standby mode. In standby mode, the components will only retain the core control module to reduce power consumption. This hardware design provides a foundation for resource-efficient sleep monitoring, that is, using only the data collected in normal collection mode for sleep staging and decreasing resource expenditure by switching to standby mode.\par
Existing studies on resource-efficient automatic sleep staging largely focus on model parameters while neglecting resource-efficient work on EEG monitors. In contrast, our work explicitly considers these factors by introducing a signal-efficient and lightweight model MASS, which leverages patch-level and epoch-level correlations through a multi-level masking and prompt learning strategy. This design enables accurate sleep staging with only partial input, making it well suited for resource-constrained environments. 

\section{Problem Formulation}
Traditional sleep staging approaches assume access to complete and continuous EEG signals over the full sleep duration, which imposes substantial demands on data acquisition, transmission, and energy resources. In this paper, we formulate a new resource-constrained sleep staging task under partial observation. Given an input EEG segment $\mathbf{X} \in \mathbb{R}^d$ corresponding to a 30-second sleep epoch, we first divide it into $T$ consecutive and non-overlapping short segments (e.g., 1-second intervals), which leads to 
$\mathbf{X} = \{ \mathbf{x}_1, \mathbf{x}_2, \dots, \mathbf{x}_T \}, \quad \mathbf{x}_t \in \mathbb{R}^{d/T}$. Instead of using all $T$ segments for classification, we assume only a subset $\mathcal{T}_{\text{obs}} \subset \{1, \dots, T\}$ of segments is available for observation and modeling. The partially observed input becomes:$\mathbf{X}_{\text{obs}} = \{ \mathbf{x}_t \mid t \in \mathcal{T}_{\text{obs}} \}$. Given the observed partial EEG sequence $\mathbf{X}_{\text{obs}}$, the goal is to learn a mapping function:$h: \mathbf{X}_{\text{obs}} \mapsto y$, where $y \in \mathcal{Y}$ denotes the sleep stage label associated with the full 30-second EEG segment, and $\mathcal{Y}$ is the set of predefined sleep stages (e.g., Wake, N1, N2, N3, REM).\par

\section{Methods}
\subsection{Overview}
\begin{figure*}[!htbp]
    \centering
    \includegraphics[width=\linewidth]{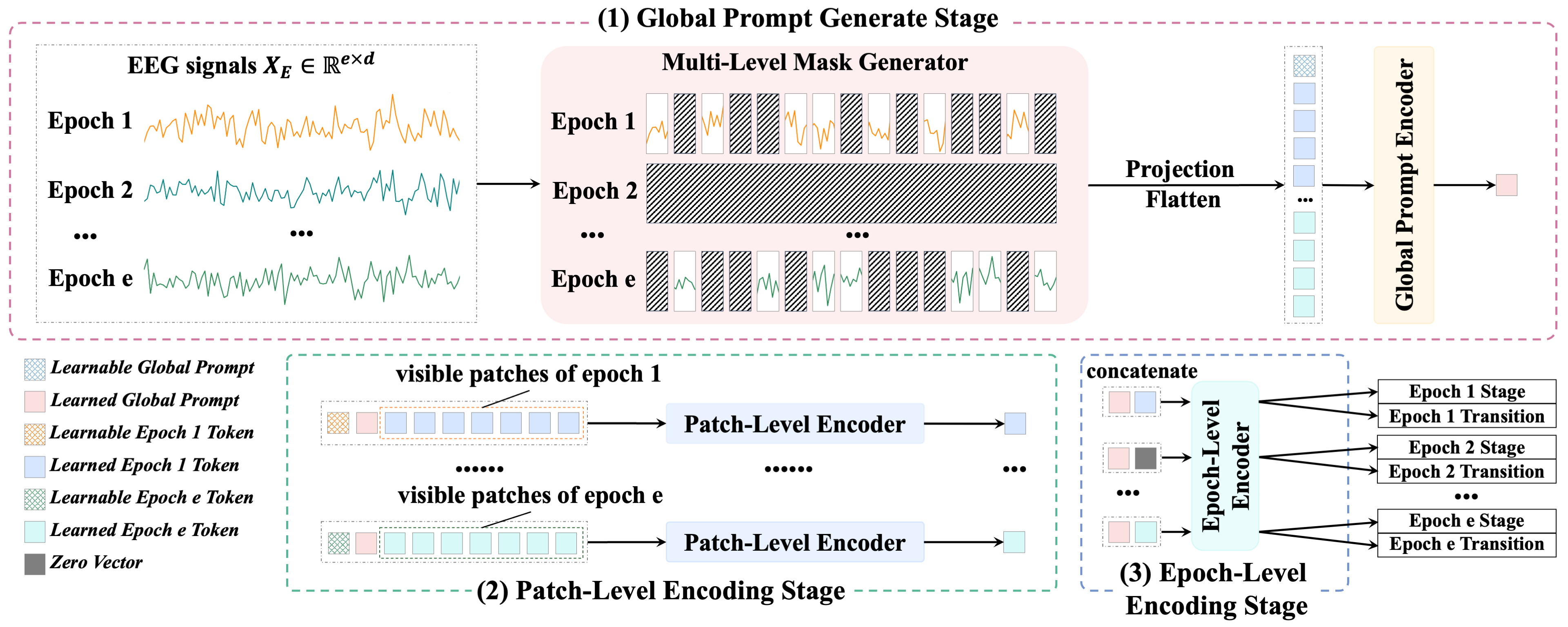}
    \caption{Complete Structure of the Proposed Mask-Aware Sleep Staging Framework.}
    \label{fig:whole}
\end{figure*}
Our novel Mask-Aware Sleep Staging (MASS) framework is illustrated as Fig.~\ref{fig:whole}. In the training phase of MASS, the input EEG signals $X_E\in \mathbb R^{e\times d}$ represents signals from $e$ consecutive sleep epochs, where each sleep epoch contains $d$ data points. MASS first applies a mask generator to divide each epoch into patches, and then performs multi-level masking by randomly masking both entire epochs and patches within unmasked epochs. The model is trained to predict the sleep stages of all epochs based on the partially observed EEG signals. During testing, MASS directly accepts partial signals as input and outputs the predicted sleep stages.
\subsection{Multi-Level Masking}
To enable reliable sleep staging under partially observed signals, we design a multi-level mask generator that operates at both the epoch and patch levels. Given the complete input EEG data $X_E \in \mathbb{R}^{e \times d}$, where $e$ denotes the number of consecutive 30-second epochs and $d$ is the number of data points per epoch. A predefined epoch-level mask ratio $r_e$ is applied to randomly mask entire consecutive sleep epochs, retaining $e \cdot (1 - r_e)$ unmasked epochs. Each unmasked epoch is then divided into 30 non-overlapping temporal patches using a fixed 1-second window. Next, each 1-second patch is transformed into the frequency domain via power spectral density (PSD) computation, yielding clearer spectral features. Another predefined patch-level mask ratio $r_a$ is then used to randomly mask patches within each retained epoch, resulting in $30 \cdot (1 - r_a)$ unmasked patches per epoch. The remaining visible patches are passed through a linear projection layer to obtain patch-level representations. The final masked EEG representation is denoted as:
$$E_{\text{vis}} \in \mathbb{R}^{e(1 - r_e)\times 30(1 - r_a) \times d_a},$$
where $d_a$ is the feature dimension after spectral transformation and linear projection. This multi-level masking mechanism is only applied during training and operates on the complete raw EEG input to simulate partial observation. Importantly, the masking pattern is independently and randomly sampled for each training instance within a batch, thereby introducing diverse learning signals and enhancing the model's robustness to various signal loss scenarios.
\subsection{Global Prompt Learning}
To compensate for the loss of contextual information caused by partial observation, we introduce a global prompt token learning module that encodes visible EEG segments into a compact representation. Inspired by Zhao et al.~\cite{zhao2024multiscale}, the global prompt is designed to serve as a semantic prior that guides in downstream patch-level and epoch-level modeling. Unlike naive attention pooling, our prompt is constructed through a shallow Transformer that operates on the entire visible sequence, with absolute positional encoding to preserve the location of each retained patch in the full EEG sequence. This enables the prompt to explicitly learn the overall context under the masking strategy. After multi-level masking, we gather all visible EEG patches across the retained epochs into a flattened sequence $E_{\text{vis}} = \{ \mathbf{e}_1, \mathbf{e}_2, \dots, \mathbf{e}_{N_{\text{vis}}} \} \in \mathbb{R}^{N_{\text{vis}} \times d_a}$, where $N_{\text{vis}} = \lfloor e \cdot (1 - r_e) \cdot 30 \cdot (1 - r_a) \rfloor$. Here, each $\mathbf{e}_i \in \mathbb{R}^{d_a}$ is a spectral-domain representation of a 1-second patch after linear projection. These patches are directly flattened across all epochs, preserving their global order in the original full sequence of $e \times 30$ patches.\par
We utilize a learnable CLS token $\mathbf{z}_0 \in \mathbb{R}^{1 \times d_a}$ to serve as the learnable global prompt $\hat{E}_{\text{vis}} = [\mathbf{z}_0; \mathbf{e}_1, \dots, \mathbf{e}_N] \in \mathbb{R}^{(N+1) \times d_a}$. The CLS token is assigned at position index 0. To maintain temporal consistency, we adopt fixed sinusoidal positional encoding, assigning each patch its position $p_i \in \{0, 1, ..., e \cdot 30 - 1\}$ from the origin full sequence before masking. The positional encoding $\mathbf{P} \in \mathbb{R}^{(N+1) \times d_a}$ is computed as:
\begin{align*}
\mathbf{P}_{p, 2i} &= \sin\left(\frac{p}{10000^{2i/d_a}}\right), \\
\mathbf{P}_{p, 2i+1} &= \cos\left(\frac{p}{10000^{2i/d_a}}\right),
\end{align*}

where the final position-aware sequence is $\tilde{E}_{\text{vis}}=\hat{E}_{\text{vis}}+P$. This position-aware sequence is then processed by a shallow Transformer encoder $f_{\text{prompt}}$ with $L_p$ layers:
$$\mathbf{Z}^{(l)} = \text{TransformerLayer}^{(l)}(\mathbf{Z}^{(l-1)}), \quad \mathbf{Z}^{(0)} = \tilde{E}_{\text{vis}}$$
The final global prompt token is extracted from the output corresponding to the CLS token as $z_{\text{prompt}} = \mathbf{Z}^{(L_p)}_0 \in \mathbb{R}^{1 \times d_a}$. This prompt summarizes all retained EEG patches using their global temporal positions and will later be injected into patch-level and epoch-level modeling modules to guide learning under partial observation.

\subsection{Patch-Level and Epoch-Level Modeling}
To effectively recognize sleep stages under partial observation, we design a two-level hierarchical modeling framework that captures both local intra-epoch dynamics and global inter-epoch transitions. The patch-level module focuses on learning fine-grained patterns from partially retained EEG patches within each epoch, while the epoch-level module aggregates semantic representations across the temporal sequence. This design ensures that both local and global temporal dependencies are preserved even when input signals are limited.\par
For each visible epoch $i$, we denote the corresponding set of unmasked EEG patches as $X_i = \{ \mathbf{x}_{i,1}, \dots, \mathbf{x}_{i,K} \} \in \mathbb{R}^{K \times d_a}$, where $K = 30 \cdot (1 - r_a)$ is the number of retained patches in the epoch, and each patch $\mathbf{x}_{i,j} \in \mathbb{R}^{d_a}$ is a spectral-domain feature. To model patch-level dynamics under partial observation and the guidance of global context, we inject the previously obtained global prompt token $z_{\text{prompt},i} \in \mathbb{R}^{1 \times d_a}$ into each epoch sequence. To ensure that the information of patch-level learning can be utilized by subsequent epoch-level modeling, we additionally add a learnable CLS token $\mathbf{z}_i^{\text{cls}} \in \mathbb{R}^{1 \times d_a}$ at the beginning of the input sequence to capture the semantic representation throughout the entire epoch. The full sequence for patch-level encoding becomes: $$\hat{X}_i = [\mathbf{z}_i^{\text{cls}}; z_{\text{prompt}}; \mathbf{x}_{i,1}, \dots, \mathbf{x}_{i,K}] \in \mathbb{R}^{(K+2) \times d_a}$$
To preserve temporal order within the epoch, we apply fixed sinusoidal positional encoding to the patch tokens according to their local indices within each epoch as $\mathbf{H}_i^{(0)} = \hat{X}_i + \text{PatchPos}$. The resulting position-aware sequence is passed through a Transformer encoder $f_{\text{patch}}$ with $L_{a}$ layers:$$\mathbf{H}_i^{(l)} = \text{TransformerLayer}^{(l)}(\mathbf{H}_i^{(l-1)})$$
The output corresponding to the CLS token is taken as the feature representation for epoch $i$ as $\mathbf{h}_i^{\text{patch}} = \left(\mathbf{H}_i^{(L_{a})}\right)_0 \in \mathbb{R}^{1 \times d_a}$, which encodes both local intra-epoch patterns and global contextual priors from the prompt token.\par
Sleep epochs are highly correlated, and changes in sleep stages usually follow specific transition patterns and temporal continuity. To model this sequential dependency, we perform epoch-level modeling across the full sequence of $e$ epochs, including both visible and masked ones. For masked epochs, we insert zero vectors as $\mathbf{h}_i^{\text{patch}}=0 \in \mathbb{R}^{1 \times d_a}$ to maintain temporal alignment. For visible epochs, we use the features $\mathbf{h}_i^{\text{patch}} \in \mathbb{R}^{1 \times d_a}$ obtained from patch-level modeling. To provide global contextual guidance, we concatenate the global prompt token $z_{\text{prompt}} \in \mathbb{R}^{1 \times d_a}$ with each epoch-level feature, forming the input as $\tilde{\mathbf{h}}_i = [\mathbf{h}_i^{\text{patch}} \, || \, z_{\text{prompt}}] \in \mathbb{R}^{1 \times 2d_a}$, and all $e$ representations are stacked into a sequence as $H_{\text{seq}} = [\tilde{\mathbf{h}}_1, \tilde{\mathbf{h}}_2, \dots, \tilde{\mathbf{h}}_e] \in \mathbb{R}^{e \times 2d_a}$. We then apply a two-layer Bi-directional GRU to model the contextual transitions across epochs. The Bi-GRU processes this sequence in both temporal directions. At each timestep $t$, the forward and backward hidden states are computed as:
$$
\overrightarrow{\mathbf{h}}_t = \overrightarrow{\text{GRU}}(\overrightarrow{\mathbf{h}}_{t-1}, \tilde{\mathbf{h}}_t), \overleftarrow{\mathbf{h}}_t = \overleftarrow{\text{GRU}}(\overleftarrow{\mathbf{h}}_{t+1}, \tilde{\mathbf{h}}_t)
$$
These two states are concatenated to form the final contextual representation of each epoch: $\mathbf{g}_t = \overrightarrow{\mathbf{h}}_t \Vert \overleftarrow{\mathbf{h}}_t \in \mathbb{R}^{2d_e}$, where $d_e$ is the dimension of hidden states in GRU kernel. By decoupling fine-grained intra-epoch encoding and coarse-grained inter-epoch temporal modeling, this two-level design enables our model to learn sleep-relevant patterns under various masking settings. The global prompt token acts as a shared semantic bridge across both levels, guiding representation learning and maintaining global coherence.

\subsection{Training}

Finally, for each epoch $i$, we employ a linear classification head to predict its sleep stage label based on the Bi-GRU representation, denoted as $\hat{y}_i = \text{MLP}(\mathbf{g}_i) \in \mathbb{R}^5$. To improve model robustness under partially observed signals, we further introduce an auxiliary stage transition prediction task to enhance the capture of inter-epoch dynamics\cite{phyo2022transsleep}. The stage transition task is formulated as a binary classification problem, where each epoch $i$ is labeled as either transitional $y_t^i = 1$ or stable $y_t^i = 0$ based on its neighbors. Given a sequence of ground-truth sleep stages \( y_s \in \mathbb{R}^e \), the transition label is defined as:
\begin{equation}
y_t^i = 
\begin{cases}
0, & \text{if } y_s^{i-1} = y_s^i = y_s^{i+1}, \\
1, & \text{otherwise}.
\end{cases}
\end{equation}
This auxiliary prediction is computed from the same GRU-derived representation $\mathbf{g}_i$ through an additional binary classification head. During training, we optimize the model using three loss components: The cross-entropy loss and the cosine similarity loss for sleep staging over 5 categories, and a binary cross-entropy loss for sleep stage transition prediction. The overall training objective is:
\begin{equation}
\mathcal{L}_{\text{total}} = \mathcal{L}_{\text{CE}} + \lambda_1 \mathcal{L}_{\text{Cos}} + \lambda_2 \mathcal{L}_{\text{Trans}},
\end{equation}
where \( \lambda_1 \) and \( \lambda_2 \) are hyperparameters controlling the balance among losses. This joint optimization enables the model to perform stage classification while remaining transition-aware, which is particularly beneficial under partially masked conditions.

\section{Experiments}
\subsection{Datasets and Settings}
We evaluated the performance of MASS on four different public sleep staging datasets: \textbf{I. DREAMS-SUB} contains 20 overnight sleep records from 20 healthy subjects aged from 20-65 years~\cite{devuyst2005dreams}. \textbf{II. Sleep-EDF-20} is from PhysioBank, which contains 39 overnight sleep records from 20 subjects~\cite{kemp2000analysis}. \textbf{III. Sleep-EDF-78} is also from PhysioBank, which contains 153 overnight sleep records from 78 subjects~\cite{kemp2000analysis}. \textbf{IV. SHHS} contains 329 overnight sleep records from 329 selected subjects~\cite{zhang2018national}~\cite{quan1997sleep}~\cite{li2022deep}. To maintain consistency with previous work, we selected the same Cz-A1 channel on DREAMS-SUB dataset~\cite{zhang2023shnn}, Fpz-Cz channel on Sleep-EDF-20 and Sleep-EDF-78~\cite{eldele2021attention}, and C4-A1 channel on SHHS to ensure a fair comparison~\cite{eldele2021attention}. To ensure the reliability of the experiment, we performed 20-fold cross-validation on all datasets. Specifically, on the DREAMS-SUB and Sleep-EDF-20 datasets, we directly divided each subject into a group. On the Sleep-EDF-78 and SHHS datasets, we randomly assigned subjects to 20 groups. For each training session, we selected 19 groups as the training set and left one group as the test set, repeating this process 20 times and averaging the accuracy and macro-F1 results.\par
We implemented MASS based on NVIDIA RTX 4090 GPU, PyTorch 2.5.1, Python 3.12, CUDA 12.4, and the source code is publicly available. MASS is optimized by Lion, the learning rate is set to 1e-4 and the weight decay is set to 1e-2. The consecutive sleep epoch number $e$ is set to 32. The global prompt encoder layer $L_p$ is set to 4. For the patch-level modeling, the patch dimension $d_a$ is set to 128 and the encoder layer $L_{a}$ is set to 4. For the Epoch-level modeling, the hidden states in Bi-GRU $d_e$ is set to 256. For the total loss function, the weight coefficient $\lambda_1$ is set to 2 and $\lambda_2$ is set to 0.5.
\subsection{Results and Analysis}
\subsubsection{Comparison on Resource-Efficient Sleep Staging}
\begin{table*}[!htbp]
\renewcommand{\arraystretch}{0.9}
\centering
\begin{adjustbox}{max width=\linewidth}
\begin{tabular}{cc|cc|cc|cc|cc} 
\toprule 
\multirow{2}{*}{Datasets}&\multirow{2}{*}{Method}&\multicolumn{2}{c|}{Full Signal}&\multicolumn{2}{c|}{72\% Signal}&\multicolumn{2}{c|}{40\% Signal}&\multicolumn{2}{c}{10\% Signal} \\
\multirow{2}{*}{}&\multirow{2}{*}{}&ACC(\%)&macro-F1(\%)&ACC(\%)&macro-F1(\%)&ACC(\%)&macro-F1(\%)&ACC(\%)&macro-F1(\%) \\
\midrule 
\multirow{8}{*}{DREAMS-SUB}&DeepSleepNet (2017)&76.11&67.02&22.31&14.42&21.80&13.37&23.51&11.08 \\
\multirow{7}{*}{}&TinySleepNet (2020)&84.83&78.49&69.78&62.48&51.56&42.64&19.30&8.81 \\
\multirow{7}{*}{}&AttnSleep (2021)&82.39&74.48&69.02&59.36&40.30&27.53&17.81&5.99 \\
\multirow{7}{*}{}&LGSleepNet (2023)&82.47&74.31&67.58&59.20&41.40&31.36&17.76&6.62 \\
\multirow{7}{*}{}&\makecell{CNN-Transformer\\-LSTM (2023)}&71.02&55.23&61.53&47.57&43.29&30.91&20.37&10.76 \\
\multirow{7}{*}{}&NeuroNet (2024)&86.36&79.51&64.77&54.63&39.42&27.47&18.90&6.69 \\
\multirow{7}{*}{}&\textbf{MASS (Ours)}&\textbf{86.76}&\textbf{81.14}&\textbf{86.76}&\textbf{80.71}&\textbf{86.35}&\textbf{80.02}&\textbf{83.31}&\textbf{75.58} \\
\midrule 
\multirow{8}{*}{Sleep-EDF-20}&DeepSleepNet (2017)&81.75&76.65&20.18&13.86&20.17&12.77&19.92&10.27 \\
\multirow{7}{*}{}&TinySleepNet (2020)&84.71&78.41&70.48&65.17&52.53&44.12&13.11&9.85 \\
\multirow{7}{*}{}&AttnSleep (2021)&84.21&77.92&72.62&63.39&56.93&44.15&18.62&16.33 \\
\multirow{7}{*}{}&LGSleepNet (2023)&83.91&77.36&69.83&58.65&57.38&43.12&31.83&23.78 \\
\multirow{7}{*}{}&\makecell{CNN-Transformer\\-LSTM (2023)}&76.66&62.55&68.44&54.94&43.50&29.33&20.49&7.97 \\
\multirow{7}{*}{}&NeuroNet (2024)&85.80&78.65&66.51&57.05&49.05&33.70&15.12&8.94 \\
\multirow{7}{*}{}&\textbf{MASS (Ours)}&\textbf{85.93}&\textbf{80.11}&\textbf{85.66}&\textbf{79.26}&\textbf{85.80}&\textbf{79.21}&\textbf{83.81}&\textbf{76.62} \\
\midrule 
\multirow{8}{*}{Sleep-EDF-78}&DeepSleepNet (2017)&77.81&71.81&31.07&23.45&29.07&19.59&26.22&13.97 \\
\multirow{7}{*}{}&TinySleepNet (2020)&79.57&74.11&73.18&68.25&63.32&52.49&39.39&26.87 \\
\multirow{7}{*}{}&AttnSleep (2021)&78.47&73.90&70.68&60.81&61.87&45.15&42.39&24.72 \\
\multirow{7}{*}{}&LGSleepNet (2023)&78.03&73.58&55.72&49.40&49.48&37.89&34.84&30.42 \\
\multirow{7}{*}{}&\makecell{CNN-Transformer\\-LSTM (2023)}&75.90&68.67&65.47&60.17&40.94&32.13&25.17&14.32 \\
\multirow{7}{*}{}&NeuroNet (2024)&\textbf{81.06}&\textbf{75.73}&69.44&60.23&57.08&40.40&35.13&20.18 \\
\multirow{7}{*}{}&\textbf{MASS (Ours)}&80.34&75.02&\textbf{80.39}&\textbf{74.95}&\textbf{80.08}&\textbf{74.42}&\textbf{77.90}&\textbf{71.61}\\
\midrule 
\multirow{8}{*}{SHHS}&DeepSleepNet (2017)&81.01&73.91&33.48&29.99&27.20&22.44&21.69&12.88 \\
\multirow{7}{*}{}&TinySleepNet (2020)&83.30&75.22&74.10&65.43&61.47&47.24&41.11&26.49 \\
\multirow{7}{*}{}&AttnSleep (2021)&81.73&72.94&70.24&57.00&57.47&36.87&22.29&10.33 \\
\multirow{7}{*}{}&LGSleepNet (2023)&82.53&73.20&71.04&57.24&56.04&37.92&26.92&15.06 \\
\multirow{7}{*}{}&\makecell{CNN-Transformer\\-LSTM (2023)}&81.71&72.76&68.35&57.90&47.72&34.80&17.42&7.89 \\
\multirow{7}{*}{}&NeuroNet (2024)&\textbf{84.94}&\textbf{76.87}&69.59&57.93&49.78&35.57&19.95&10.65 \\
\multirow{7}{*}{}&\textbf{MASS (Ours)}&84.24&\textbf{76.87}&\textbf{84.38}&\textbf{76.58}&\textbf{83.60}&\textbf{75.18}&\textbf{80.57}&\textbf{70.25} \\
\bottomrule 
\end{tabular}
\end{adjustbox}
\caption{Comparison with State-of-The-Art Models on Four Datasets.}
\label{tab:Comparison}
\end{table*}

We compared MASS with other state-of-the-art sleep staging methods including: \textbf{I. DeepSleepNet} is the traditional CNN-BiLSTM network for extracting local and transition features~\cite{supratak2017deepsleepnet}. \textbf{II. AttnSleep} utilized multi-resolution convolution kernel and multi-head attention to capture features~\cite{eldele2021attention}. \textbf{III. TinySleepNet} is a classical model based on CNN and RNN~\cite{supratak2020tinysleepnet}. \textbf{IV.CNN-Transformer-LSTM} utilizes a CNN network to capture time-invariant features and a Transformer Encoder followed by a LSTM network to capture temporal dependency relationships as well as transition rules among sleep epochs~\cite{pham2023automatic}. \textbf{V.LGSleepNet} is a hybrid neural network with deep adaptive orthogonal fusion to extract and fusion local and global features.~\cite{shen2023lgsleepnet} \textbf{VI.NeuroNet} has a pretrained encoder network followed by a Mamba-based~\cite{gu2024mamba} temporal context module, which is applied for predicting sleep stages~\cite{lee2024neuronet}. We implemented these methods based on their publicly available code and paper description. To simulate the limited resource scenario in sleep monitoring, we randomly mask the test data with different mask ratio to set four different signal integrity levels: 100\%, 72\%, 40\%, and 10\%. For the models only accepts one sleep epoch signal as input without epoch-level modeling (AttnSleep and LGSleepNet), we directly mask their input data with the corresponding mask ratio. For the rest models using multi-epoch signals as inputs, we set the patch-level mask ratio $r_a$ and epoch-level mask ratio $r_e$ as $0.0+0.0$, $0.2+0.1$, $0.5+0.2$ and $0.8+0.5$ correspondingly, maintaining consistency with MASS. We choose these settings since they cover a wide range of signal completeness, more experiment results of MASS with different mask ratio $r_a$ and $r_e$ are provided in the supplementary material. Tab.~\ref{tab:Comparison} shows the performance comparison under various signal integrities.\par
As shown in Table~\ref{tab:Comparison}, MASS consistently achieves the best or highly competitive performance across all four datasets and under all levels of signal integrity, especially when the data is highly limited. Specifically, under full signal (100\%) situation, MASS achieves the highest accuracy and macro-F1 scores on DREAMS-SUB and Sleep-EDF-20 datasets, and nearly best accuracy and macro-F1 scores on Sleep-EDF-78 and SHHS datasets. As the signal integrity decreases, the performance of existing methods degrades rapidly since they did not consider the efficient utilization of resources at the initial stage of design and training. At the 72\% signal level, MASS achieves impressive performance, which outperforms the strongest baseline by up to +18.2\%, +14.1\%, +6.7\%, and +11.2\% macro-F1 scores on four datasets, respectively. At 40\% signal integrity, MASS still maintains robust performance, which outperforms the strongest baseline by up to +37.4\%, +35.1\%, +21.9\%, and +27.9\% macro-F1 scores on four datasets, respectively. At the extreme scenario with 10\% signal, MASS still maintains high performance under different signal resource, which outperforms the strongest baseline by up to +64.5\%, +52.8\%, +41.2\%, and +43.8\% macro-F1 scores on four datasets, respectively. Compared with the performance under complete signals, the accuracy under 10\% signal only decreased 3.45\%, 2.12\%, 2.44\%, and 3.67\%, while the macro-F1 only decreased 5.56\%, 3.49\%, 3.41\%, and 6.62\% on four datasets, respectively. These results collectively demonstrate that MASS not only performs strongly under ideal full signal settings but also significantly outperforms existing state-of-the-art models in realistic low-resource scenarios, making it a highly promising solution for resource-efficient wearable sleep monitoring systems.\par

\begin{figure}[ht]
    \centering
    \includegraphics[width=\linewidth]{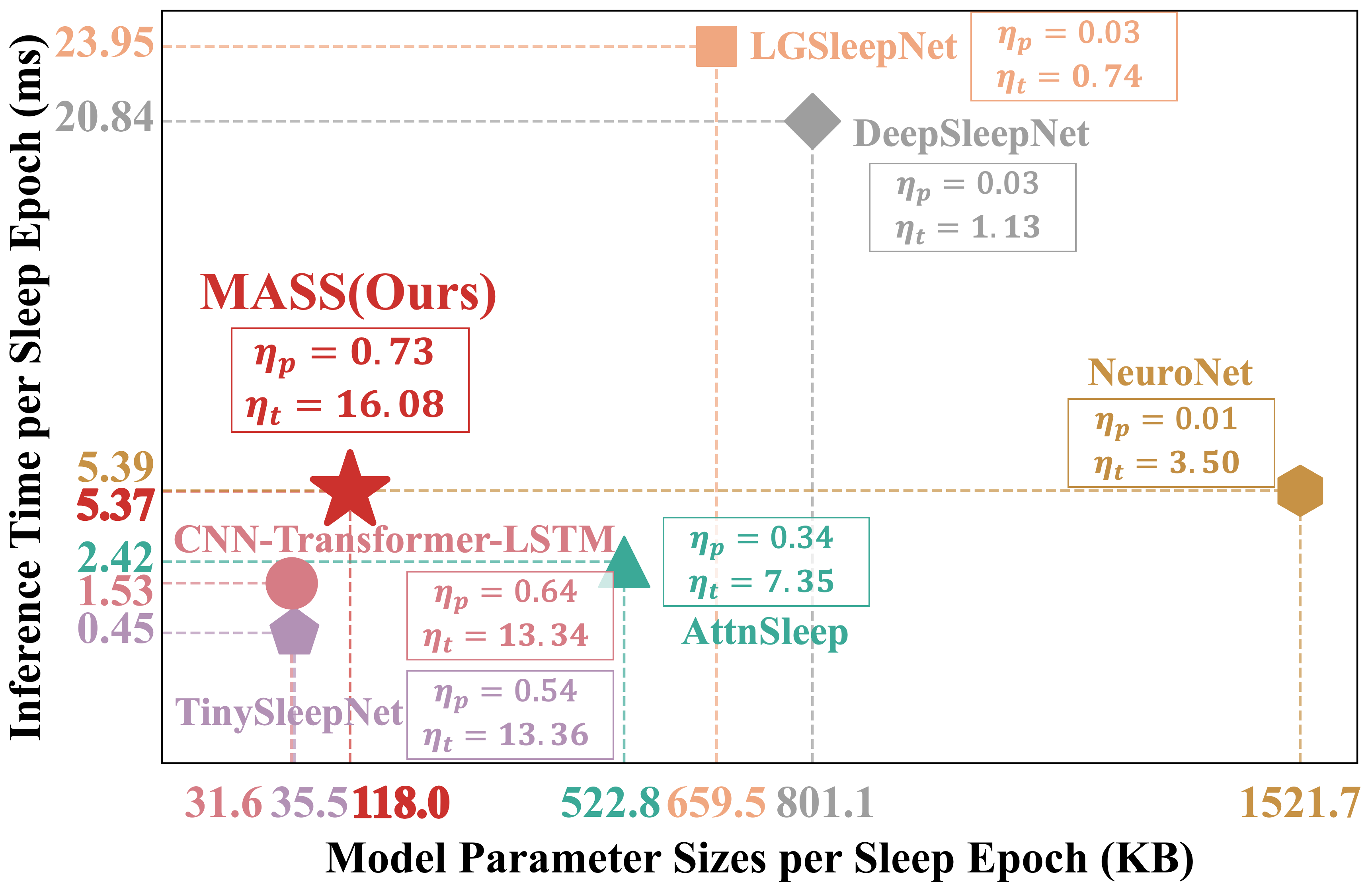}
    \caption{Comparison of Model Parameter Sizes and Inference Time on DREAMS-SUB dataset. The results are calculated in 10\% signal integrity (80\% patch-level masking and 50\% epoch-level masking). MASS achieves highest $n_p$ and $n_t$, especially compared with latest models such as LGSleepNet(2023) and NeuroNet(2024).}
    \label{fig:Params}
\end{figure}

For resource-constrained and time-sensitive tasks of sleep staging, it is important to consider the integrated factors in terms of reference time consumption, model size (i.e., the number of parameters), and the achieved accuracy. Thus, we adopt the following metrics to evaluate such integrated performance as follows:
$$\eta_p=ACC/P_{\textit{model}}, \eta_t=ACC/T_{\textit{inference}}$$
Here, $P_{\textit{model}}$ denotes the number of model parameters per sleep epoch, and $T_{\textit{inference}}$ is the inference time per sleep epoch. $\eta_p$ and $\eta_t$ reflects how effectively a model converts computational and storage resources into predictive performance. As shown in Fig.~\ref{fig:Params}, our proposed MASS model achieves the highest values in both $\eta_p=0.73$ and $\eta_t=16.08$, surpassing all other methods in both metrics. This demonstrates that MASS not only provides strong performance in accuracy under low data collection costs, but also maintains high efficiency in terms of parameter and inference time overhead, making it highly suitable for deployment in resource-limited, real-time wearable sleep monitoring systems.\par

\subsubsection{Resource Utilization Study}
To verify the capability of the proposed MASS in resource-efficient sleep monitoring, we selected three commonly used signal amplification and analog-to-digital conversion modules in current flexible wearable monitoring systems for comparison: \textbf{I. ADS1299-4 II. ADS131A04 III. ADS1294}~\cite{ADS1299}~\cite{ADS131A04}~\cite{ADS1294}. They are all small-sized, high-performance signal processors, supporting normal mode and standby mode switching within a few decimal time periods. According to the relevant descriptions in the manual, the collection costs under different signal integrity levels are calculated in Table~\ref{tab:Amp}. These results further demonstrate the efficiency and scalability of MASS for practical deployment in resource-constrained scenarios. These results indicate that, by reducing the signal acquisition ratio to as low as 10\%, the overall power consumption of mainstream amplifiers can be decreased by more than 60\%, significantly prolonging battery life and enhancing user comfort in wearable applications. Notably, even under such limited signal conditions, our proposed MASS framework still maintains high performance in sleep staging as demonstrated in Table~\ref{tab:Comparison}. This further confirms that MASS is well-suited for practical deployment in resource-constrained scenarios, achieving a desirable balance between sleep staging accuracy and hardware efficiency.

\begin{table}[b]
\centering
\begin{adjustbox}{max width=\linewidth}
\begin{tabular}{c|ccc} 
\toprule 
Amplifier&ADS1299-4&ADS131A04&ADS1294 \\
\midrule 
\makecell{Typical Power on\\Normal Mode}&22mW&15.8mW&10.1mW\\
\makecell{Typical Power on\\Standby Mode}&5.1mW&2.6mW&4mW\\
\midrule 
72\% Signal&17.27mW&12.10mW&8.39mW\\
40\% Signal&11.86mW&7.88mW&6.44mW\\
10\% Signal&6.79mW&3.92mW&4.61mW\\
\bottomrule 
\end{tabular}
\end{adjustbox}
\caption{Comparison of Data Acquisition Resource Costs of Different Amplifiers Under Different Signal Integrities.} 
\label{tab:Amp}
\end{table}

\subsubsection{Ablation Study}
To investigate the effectiveness of the components in our proposed MASS, we conducted a component ablation study. Specifically, through module reduction, we obtained the following four models: \textbf{I. MASS-Base}, which includes only the simplest patch-level and epoch-level feature learning \textbf{II. MASS-Prompt}, which adds a global prompt as semantic guidance on the basis of MASS-Base \textbf{III. MASS-Mask}, which adds multi-level masking on the basis of MASS-Base \textbf{IV. MASS}, the complete model. We conducted ablation experiments on four datasets with a unified setting of 40\% data integrity, and recorded the corresponding accuracy and macro-F1 metric results. As shown in Table~\ref{tab:Ablation}, removing the multi-level masking (MASS-Base and MASS-Prompt) leads to a significant drop in performance across all datasets, highlighting the importance of training under partial observation. The addition of global prompt tokens (MASS-Prompt) brings moderate gains over the base model, suggesting that global semantic guidance can facilitate representation learning. The full model MASS achieves the best overall performance, demonstrating that combining hierarchical modeling, prompt-based guidance, and masking strategy results in a more robust and effective architecture for sleep staging under limited signals.

\begin{table}[t]
\centering 
\begin{adjustbox}{max width=\linewidth}
\begin{tabular}{cccc} 
\toprule 
Datasets&Model&ACC(\%)&macro-F1(\%) \\
\midrule 
\multirow{4}{*}{DREAMS-SUB}&MASS-Base&43.2&15.9 \\
{}&MASS-Prompt&45.6&18.5 \\
{}&MASS-Mask&85.8&79.9 \\
{}&MASS&86.4&80.0 \\
\midrule 
\multirow{4}{*}{Sleep-EDF-20}&MASS-Base&55.4&32.7 \\
{}&MASS-Prompt&59.6&41.3 \\
{}&MASS-Mask&85.5&79.2 \\
{}&MASS&85.8&79.2 \\
\midrule 
\multirow{4}{*}{Sleep-EDF-78}&MASS-Base&50.1&33.8 \\
{}&MASS-Prompt&52.7&33.9 \\
{}&MASS-Mask&79.8&73.2 \\
{}&MASS&80.1&74.4 \\
\midrule 
\multirow{4}{*}{SHHS}&MASS-Base&58.0&43.2 \\
{}&MASS-Prompt&59.9&45.1 \\
{}&MASS-Mask&82.8&74.6 \\
{}&MASS&83.6&75.2 \\
\bottomrule 
\end{tabular}
\end{adjustbox}
\caption{Component Ablation Study on Four Datasets.}
\label{tab:Ablation}
\end{table}

\section{Conclusion}
In this paper, we propose Mask-Aware Sleep Staging (MASS), a novel framework for sleep staging under incomplete EEG signals, which jointly leverages multi-level masking and prompt learning to model both patch-level and epoch-level dynamics. By introducing a hierarchical masking mechanism and global prompt token as semantic prior, MASS is able to simulate varying degrees of signal incompleteness, enhancing representation learning under partial observation. Experiments on four public datasets demonstrate that MASS consistently outperforms state-of-the-art baselines, especially under low-resource settings, while maintaining moderate computational costs. These results highlight the effectiveness and practicality of our framework for deployment in real-world wearable sleep monitoring systems.

\section{Acknowledgments}
This work was supported by National Natural Science Foundation of China (No.62276109), Fundamental Research Funds for the Central Universities of China (ZZZX202402), Postdoctoral Fellowship Program of the China Postdoctoral Science Foundation (GZB20240244), China Postdoctoral Science Foundation (2024M761016), and Guangdong Basic and Applied Basic Research Foundation (2024A1515110155). The corresponding authors are Prof. Min Chen and Dr. Rui Wang.

\bibliography{aaai2026}

\clearpage
\section{Appendix}
\subsection{Data Preprocessing}
In this work, we focus on efficient learning from limited EEG signal by leveraging a multi-level masking representation learning strategy. To enhance the learning capacity of sparse temporal fragments (e.g., 1-second patch sampled from a 30-second sleep epoch), we transform the raw EEG signals from the time domain into the spectral domain via short-time Fourier transform (STFT), which provides a more compact and informative representation of local neural dynamics.\par
The original EEG recordings are resampled into 100 Hz using a rational polyphase filter, which facilitates alignment with STFT windowing configurations and results in 3000 time points per epoch. To extract spectral features, we apply STFT using a non-overlapping Hamming window of 1 second (100 points) and an FFT size of 256. This results in exactly 30 spectral patches per epoch, each corresponding to a 1-second time segment without overlapping and has 128 frequency points. These points represent spectral magnitudes across the 0-50 Hz range, covering the most relevant frequency content for sleep-related EEG patterns under a 100 Hz sampling rate. The results are then transformed to power spectrum density (PSD) to reduce dynamic range and stabilize training:
$$
\text{PSD} = 20 \cdot \log_{10}(|X(f, t)| + \epsilon), \quad \epsilon = 10^{-8}
$$
Compared to raw time-domain EEG signals, the spectral representation captures localized frequency patterns that are physiologically relevant for sleep staging (e.g., delta and theta activity during NREM stages), and more robust to local temporal distortions. This property is especially valuable under partial observation settings, where only a subset of time patches are available. In such cases, spectral features provide a semantically richer basis for modeling than raw signals, enabling effective patch-level learning with sparse input.

\subsection{Model Hyperparameters}
We summarize the key hyperparameters and training configurations of the proposed MASS model in the Table \ref{tab:hyper} and \ref{tab:train}. MASS adopts a moderate patch-level feature dimension of 128 and a hidden size of 256 at the epoch level. Both the and prompt-level epoch-level encoders consist of 4 Transformer layers, while the inter-epoch encoder is a Bi-GRU with 2 layers. For training, we use the Lion optimizer with a cosine learning rate scheduler and a warmup strategy, which helps stabilize optimization in the early training stage. All models are trained for 100 epochs with a weight decay of 0.01 to avoid overfitting. These settings strike a balance between performance and computational efficiency, making MASS suitable for deployment in lightweight, resource-constrained scenarios.

\begin{table}[!htbp]
\centering
\begin{adjustbox}{max width=\linewidth}
\begin{tabular}{c|c}
\hline
\textbf{Hyperparameter} & \textbf{Value} \\
\hline
patch-level feature dimension $d_a$ & 128 \\
epoch-level hidden dimension $d_e$ & 256 \\
number of heads in Multi-Head Attention & 8 \\
mlp ratio in Transformer encoder & 4 \\
dropout rate & 0.1 \\
number of patches & 30 \\
number of epoches & 32 \\
global encoder depth $L_p$ & 4 \\
intra encoder depth $L_a$ & 4 \\
inter encoder depth $L_e$ & 2 \\
cosine loss weight $\lambda_1$ & 2 \\
transition loss weight $\lambda_2$ & 0.5 \\
\hline
\end{tabular}
\end{adjustbox}
\caption{Hyperparameters and Values For MASS.}
\label{tab:hyper}
\end{table}

\begin{table}[!htbp]
\centering
\begin{adjustbox}{max width=\linewidth}
\begin{tabular}{c|c}
\hline
\textbf{Training Settings} & \textbf{Value} \\
\hline
Optimizer & Lion \\
Learning Rate & 1e-4 \\
Weight Decay & 0.01 \\
Scheduler & CosineWithWarmup \\
Total Epochs & 100 \\
Warmup Epochs & 10 \\
Minimum Learning Rate & 1e-6 \\
\hline
\end{tabular}
\end{adjustbox}
\caption{Training Settings for MASS.}
\label{tab:train}
\end{table}

\subsection{Dataset Description}
\begin{table*}[!htbp]
\centering
\begin{adjustbox}{max width=\linewidth}
\begin{tabular}{c|ccccc|c}
\toprule
\textbf{Dataset} & \textbf{W} & \textbf{N1} & \textbf{N2} & \textbf{N3} & \textbf{REM} & \textbf{Total} \\
\midrule
DREAMS-SUB       & 3559 (17.58\%) & 1480 (7.31\%)  & 8251 (40.76\%) & 3933 (19.43\%) & 3019 (14.91\%) & 20242 \\
Sleep-EDF-20       & 6558 (16.28\%) & 2766 (6.86\%)  & 17629 (43.75\%) & 5629 (13.97\%) & 7711 (19.14\%) & 40293 \\
Sleep-EDF-78       & 35788 (22.03\%) & 20342 (12.52\%) & 67661 (41.65\%) & 12894 (7.94\%) & 25771 (15.86\%) & 162456 \\
SHHS         & 45613 (14.07\%) & 10304 (3.18\%)  & 142125 (43.85\%) & 54690 (16.87\%) & 71416 (22.03\%) & 324148 \\
\bottomrule
\end{tabular}
\end{adjustbox}
\caption{Distribution of Sleep Stages Across Four Datasets.}
\label{tab:dataset_distribution}
\end{table*}
In this paper, we utilize four publicly available datasets for comparative experiments, namely DREAMS-SUB, Sleep-EDF-20, Sleep-EDF-78, and SHHS. Sleep-EDF-20 contains data files for 20 subjects, while Sleep-EDF-78 is an expanded version with 78 subjects. For these two datasets, each PSG file contains two EEG channels (Fpz-Cz, Pz-Oz) with a sampling rate of 100 Hz, one EOG channel and one chin EMG channel. Following previous studies, we utilized the single Fpz-Cz channel as the input for various models in our experiments.\par
The DREAMS-SUB contained 20 overnight sleep records from healthy subjects. The data were obtained with a digital 32-channel polygraph in a sleep laboratory of a Belgium hospital. Each PSG recording contained at least two EOG channels (P8-A1, P18-A1), three EEG channels (Cz-A1 or C3-A1, FP1-A1, and O1-A1), and one submental EMG channel with a sampling frequency of 200 Hz. We used Cz-A1 channel EEG to evaluate our model, and downsampled each EEG data record from the DREAMS-SUB dataset to 100 Hz for unify data preprocessing.\par
SHHS is a multi-center cohort study of the cardiovascular and other consequences of sleep-disordered breathing. The subjects suffer from various diseases including lung diseases, cardiovascular diseases and coronary diseases. To minimize the impact of these diseases, we followed the previous study to select subjects, who are considered to have a regular sleep. Eventually, 329 out of 6441 subjects were selected for our experiments. Notably, we selected the C4-A1 channel with a sampling rate of 125 Hz, and downsampled them into 100Hz for unify data preprocessing.\par
To balance the data categories, we selected the signal data from the first 30 minutes before the start of the first sleep epoch and the last 30 minutes after the end of the last sleep epoch as the wake stage data. We also excluded any UNKNOWN stages that don’t belong to any of the sleep stages. Finally, we merged stages N3 and N4 into one stage (N3) according AASM standard. The sleep epoch information for all datasets is shown in the Table \ref{tab:dataset_distribution}.

\subsection{Algorithm of MASS}
\begin{algorithm}[htbp]
\caption{Multi-level Masked Autoencoder for Sleep Staging (MASS)}
\begin{algorithmic}[1]
\REQUIRE EEG data $\mathbf{X} \in \mathbb{R}^{B \times E \times P \times 128}$
\ENSURE Sleep stage prediction $\hat{y}_{stage}$, transition prediction $\hat{y}_{trans}$

\STATE \textbf{Initialize} parameters: \texttt{embed\_dim, num\_heads, mlp\_ratio, dropout, ...}

\STATE Generate inter-epoch and intra-epoch masks:
\STATE \quad $\mathcal{M}_{epoch} \leftarrow \texttt{MaskGenerator}(E, r_e)$
\STATE \quad $\mathcal{M}_{patch} \leftarrow \texttt{MaskGenerator}(P, r_a)$

\STATE Extract visible patches and apply patch embedding:
\STATE \quad $\mathbf{Z}_{patch} \leftarrow \texttt{PatchEmbedding}(\mathbf{X}_{visible})$

\STATE Concatenate global CLS token and add global positional encoding:
\STATE \quad $\mathbf{Z}_{global} \leftarrow [\texttt{CLS}_{global}; \mathbf{Z}_{patch}] + \texttt{PosEnc}_{global}$

\STATE Compute prompt token via global Transformer:
\STATE \quad $\mathbf{H}_{global} \leftarrow \texttt{Transformer}_{global}(\mathbf{Z}_{global})$
\STATE \quad $\mathbf{p} \leftarrow \mathbf{H}_{global}[:, 0, :]$

\FOR{each visible epoch}
    \STATE Concatenate CLS, prompt, and patch tokens:
    \STATE \quad $\mathbf{Z}_{patch} \leftarrow [\texttt{CLS}_{epoch}; \mathbf{p}; \mathbf{Z}_{patch}] + \texttt{PosEnc}$
    \STATE \quad $\mathbf{H}_{patch} \leftarrow \texttt{Transformer}_{patch}(\mathbf{Z}_{patch})$
    \STATE \quad $\mathbf{c}_{epoch} \leftarrow \mathbf{H}_{patch}[:, 0, :]$
\ENDFOR

\STATE Concatenate prompt $\mathbf{p}$ and $\mathbf{c}_{epoch}$ for all epochs:
\STATE \quad $\mathbf{Z}_{epoch} \leftarrow [\mathbf{p}; \mathbf{c}_{epoch}]$

\STATE \texttt{BiGRU} modeling:
\STATE \quad $\mathbf{H}_{epoch} \leftarrow \texttt{BiGRU}(\mathbf{Z}_{epoch})$

\STATE Predict sleep stage: $\hat{y}_{stage} \leftarrow \texttt{MLP}_{stage}(\mathbf{H}_{epoch})$
\STATE Predict sleep epoch transition label: $\hat{y}_{trans} \leftarrow \texttt{MLP}_{trans}(\mathbf{H}_{epoch}[\text{masked}])$

\RETURN $\hat{y}_{stage}, \hat{y}_{trans}$
\end{algorithmic}
\end{algorithm}

\subsection{Confusion Matrix of MASS}
\begin{figure*}[htbp]
  \centering
  \begin{tabular}{cccc}
    \begin{subfigure}[b]{0.22\textwidth}
      \includegraphics[width=\linewidth]{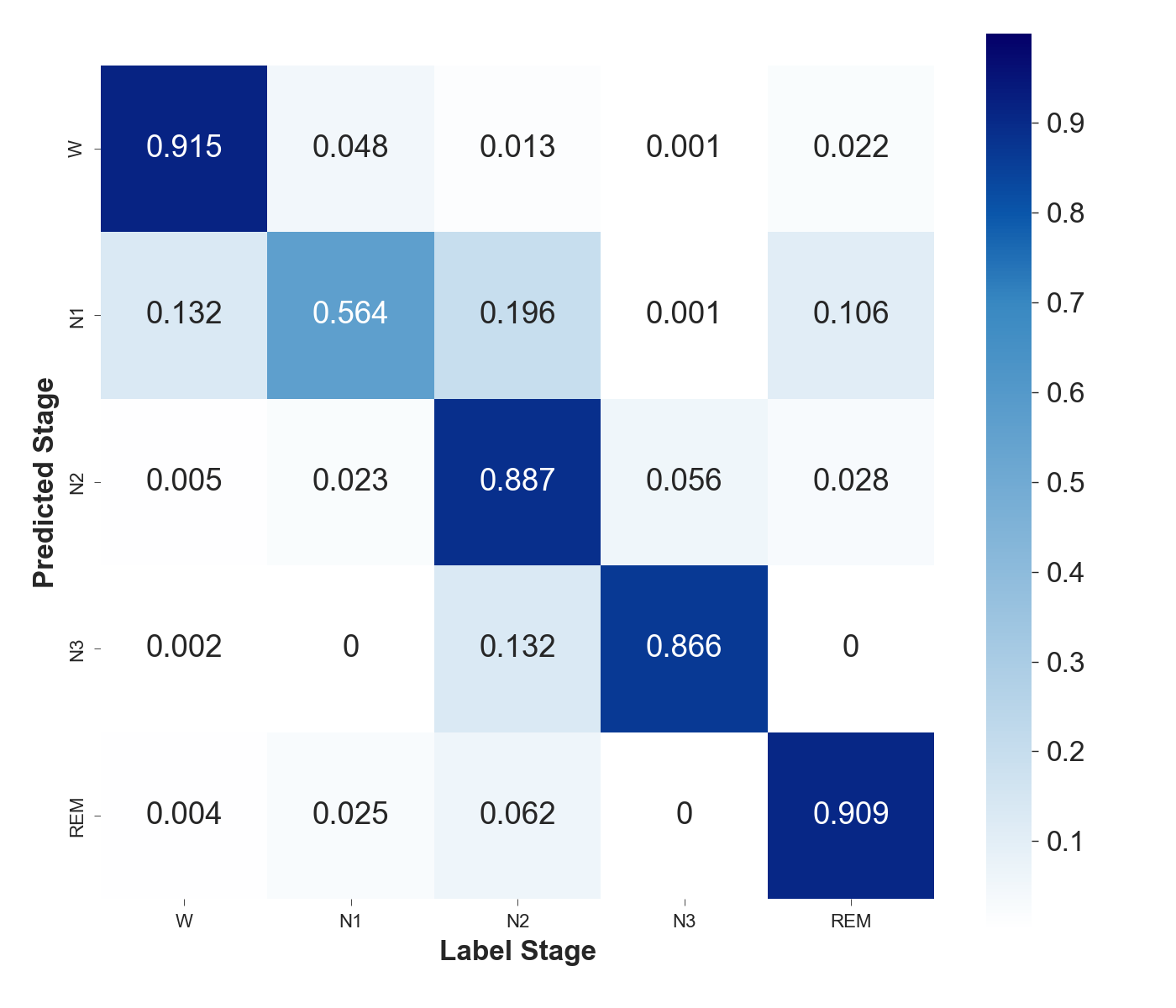}
      \caption{0\% Mask on DREAMS-SUB.}
    \end{subfigure} &
    \begin{subfigure}[b]{0.22\textwidth}
      \includegraphics[width=\linewidth]{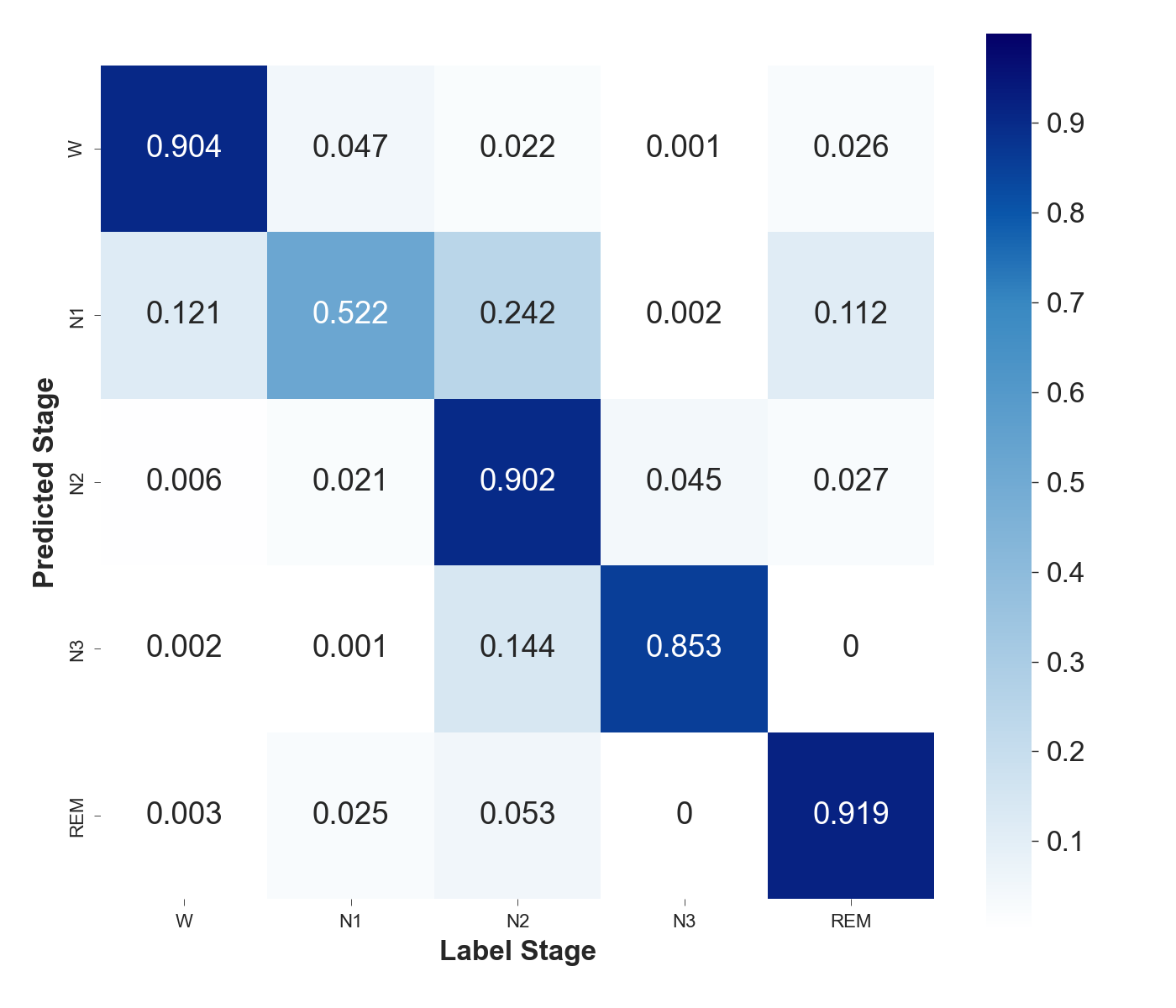}
      \caption{28\% Mask on DREAMS-SUB.}
    \end{subfigure} &
    \begin{subfigure}[b]{0.22\textwidth}
      \includegraphics[width=\linewidth]{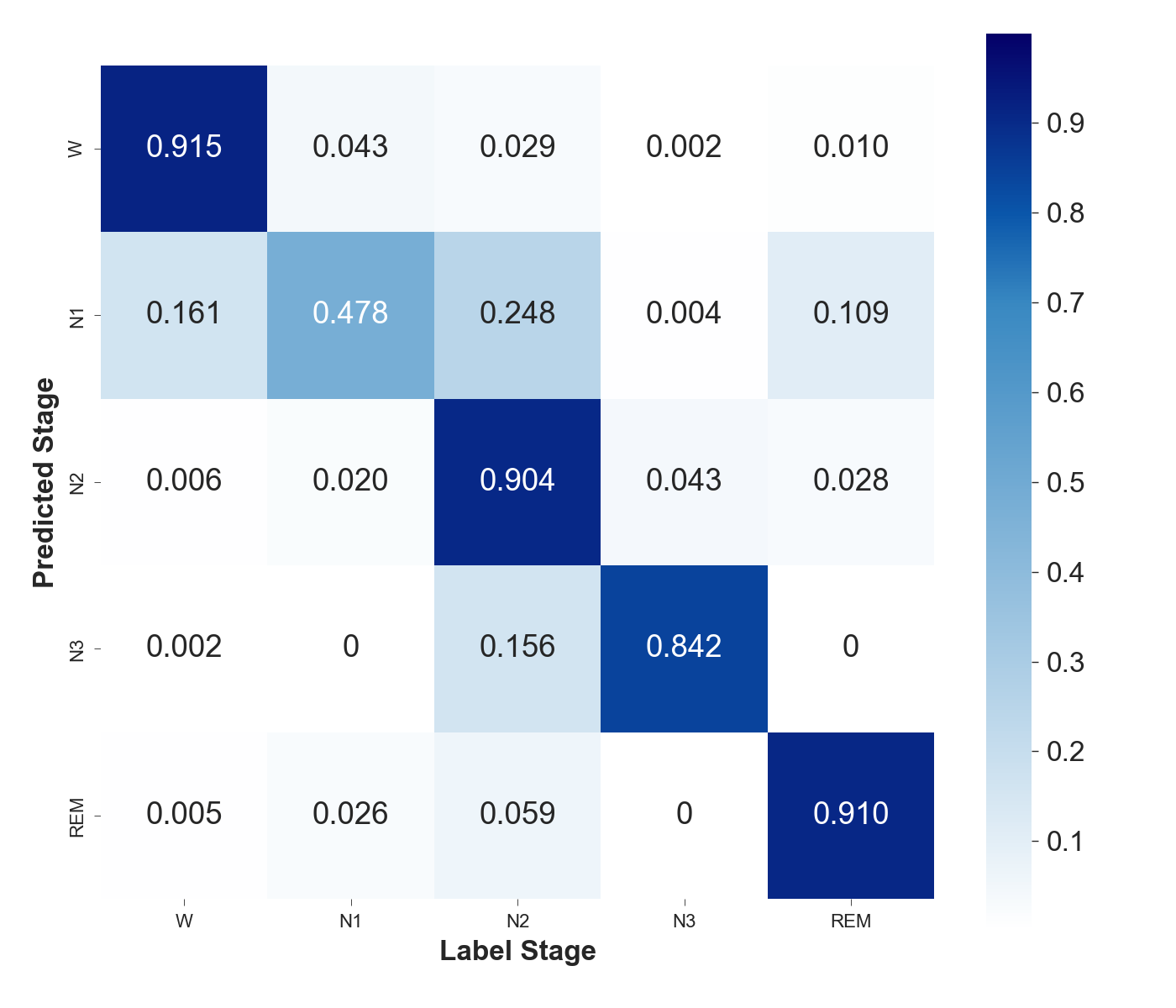}
      \caption{60\% Mask on DREAMS-SUB.}
    \end{subfigure} &
    \begin{subfigure}[b]{0.22\textwidth}
      \includegraphics[width=\linewidth]{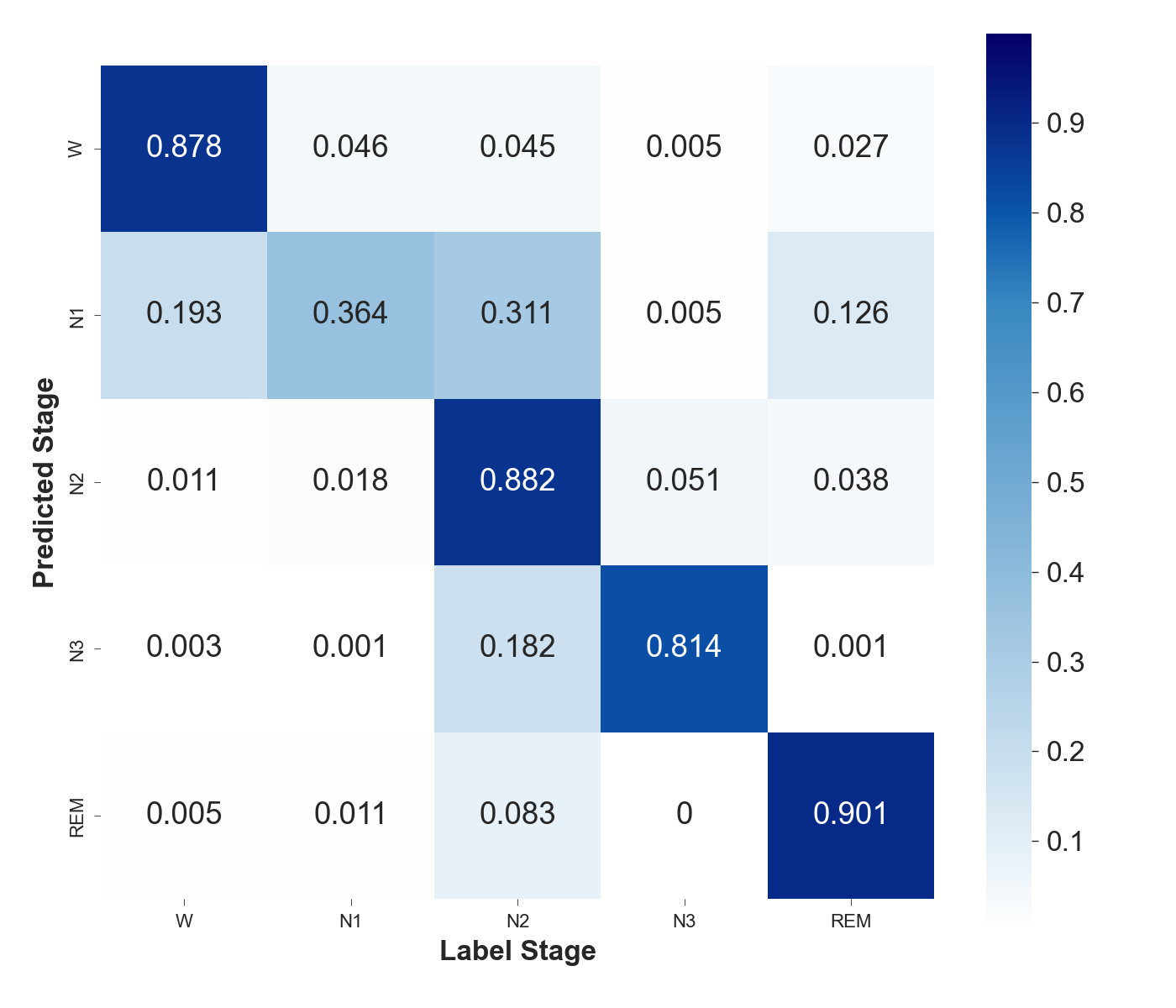}
      \caption{90\% Mask on DREAMS-SUB.}
    \end{subfigure} \\

    \begin{subfigure}[b]{0.22\textwidth}
      \includegraphics[width=\linewidth]{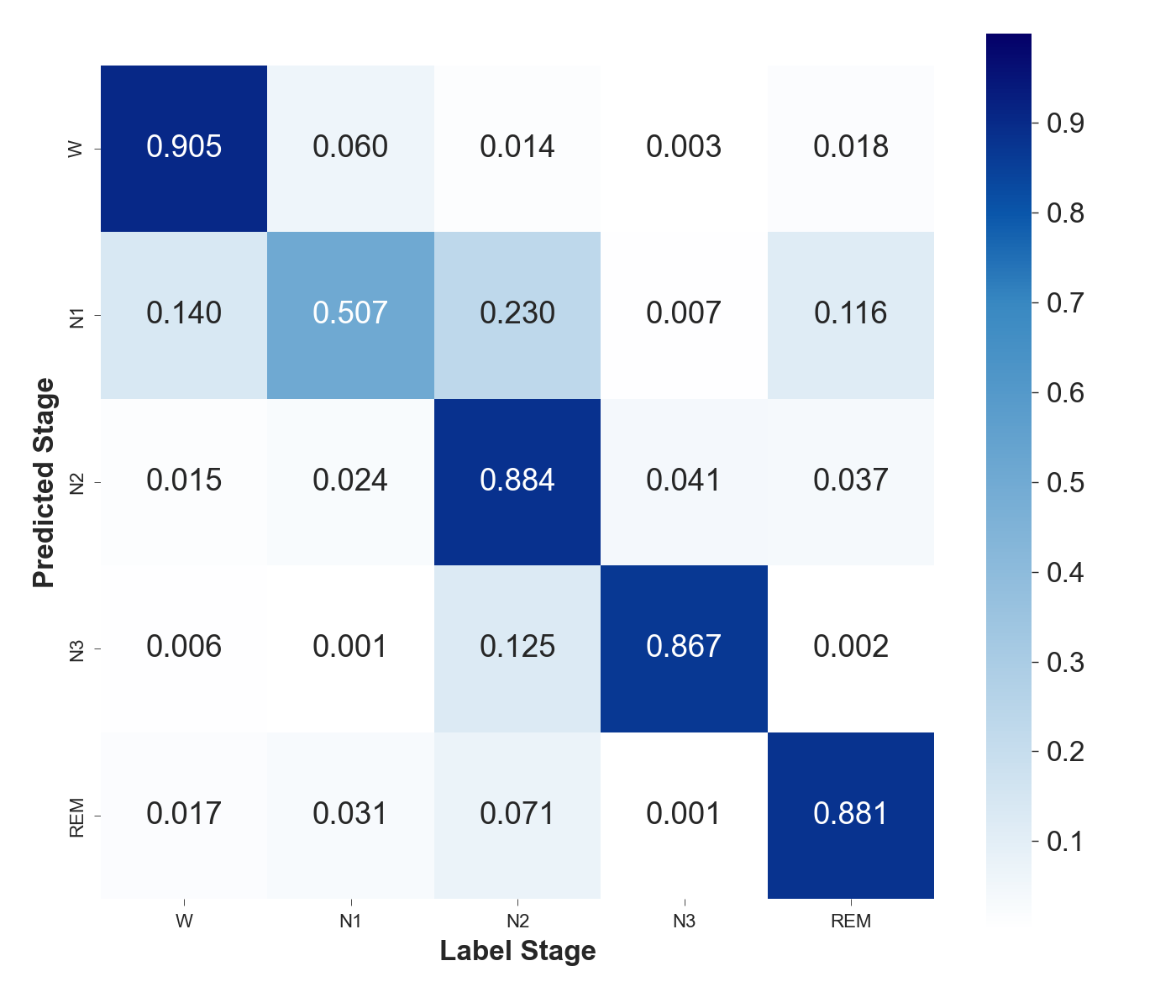}
      \caption{0\% Mask on Sleep-EDF-20.}
    \end{subfigure} &
    \begin{subfigure}[b]{0.22\textwidth}
      \includegraphics[width=\linewidth]{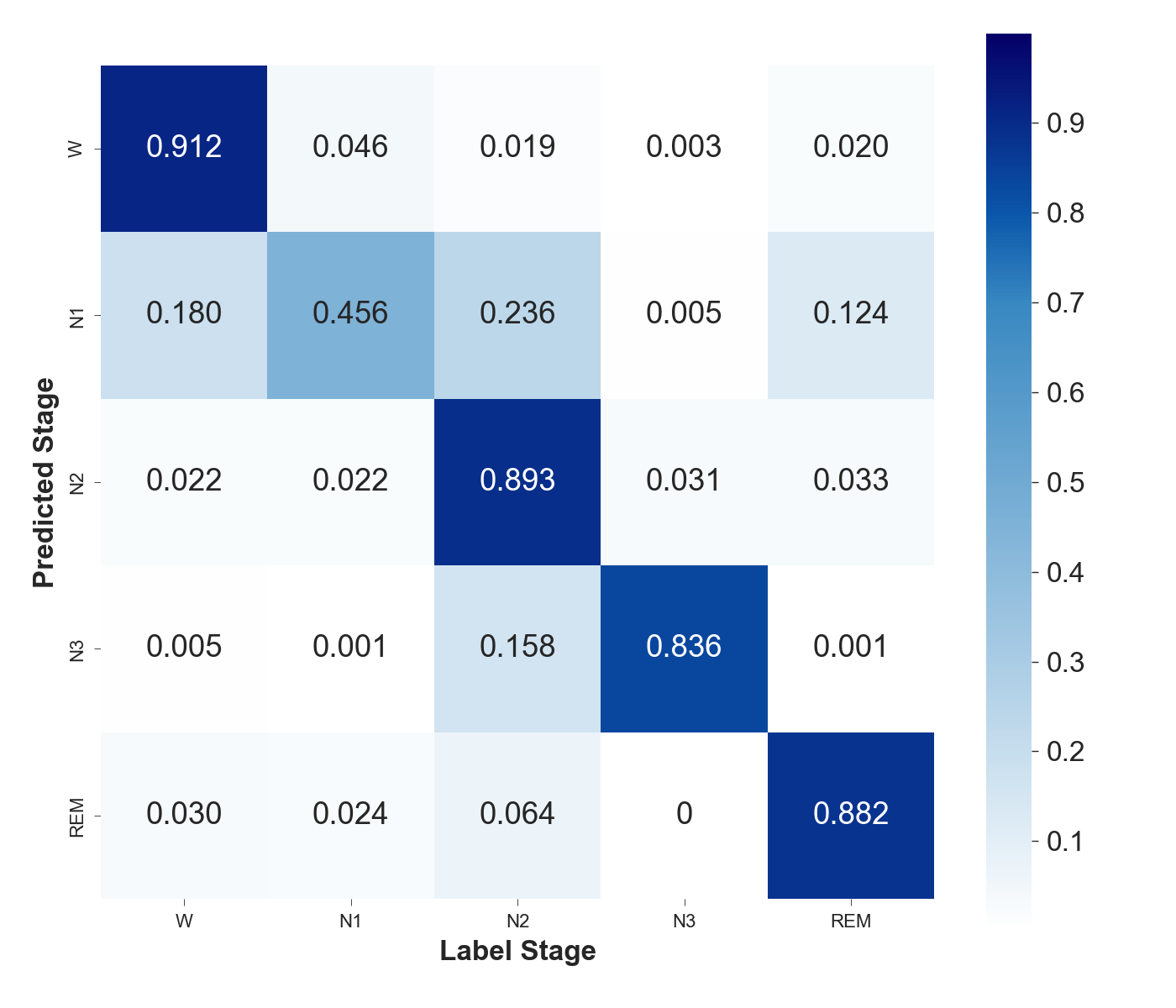}
      \caption{28\% Mask on Sleep-EDF-20.}
    \end{subfigure} &
    \begin{subfigure}[b]{0.22\textwidth}
      \includegraphics[width=\linewidth]{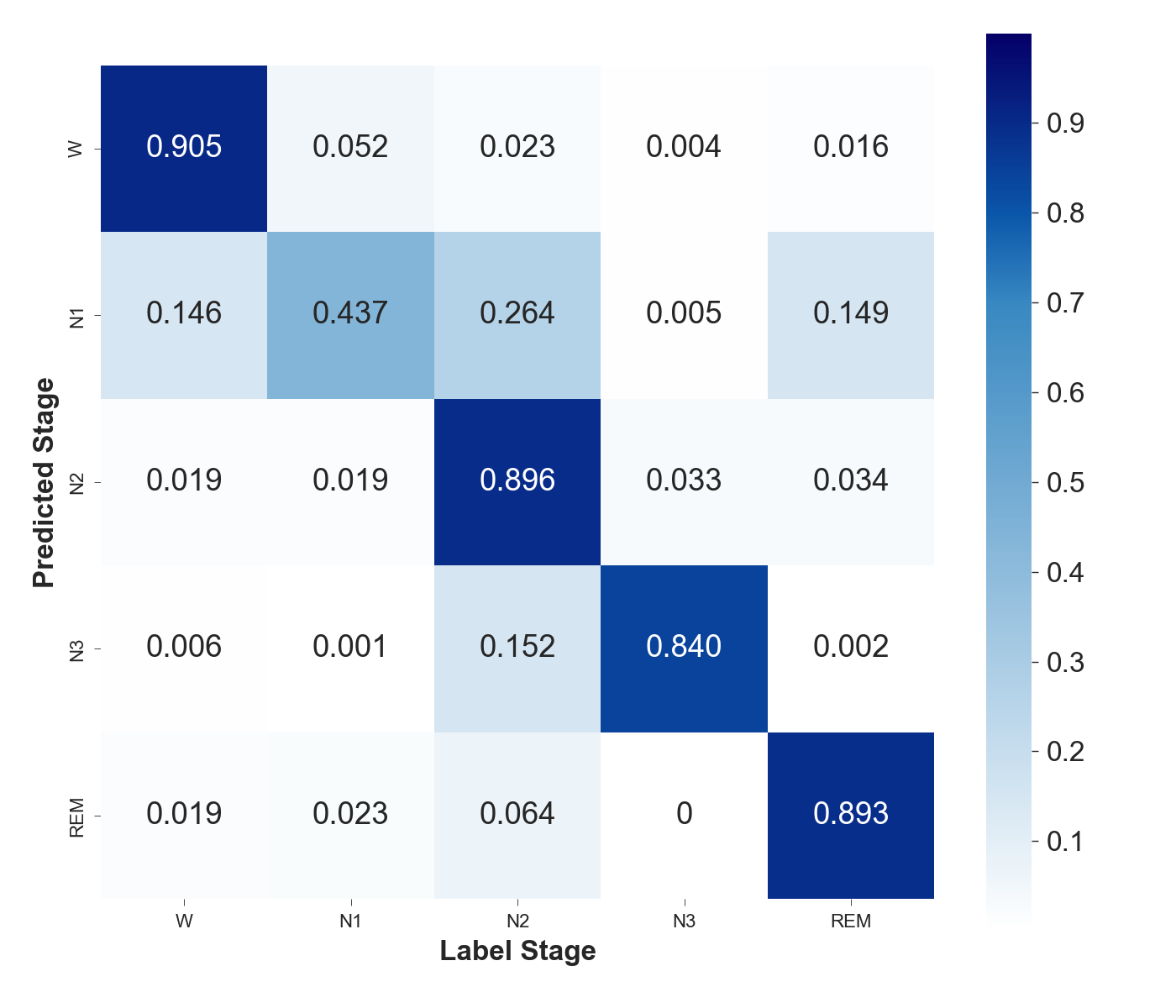}
      \caption{60\% Mask on Sleep-EDF-20.}
    \end{subfigure} &
    \begin{subfigure}[b]{0.22\textwidth}
      \includegraphics[width=\linewidth]{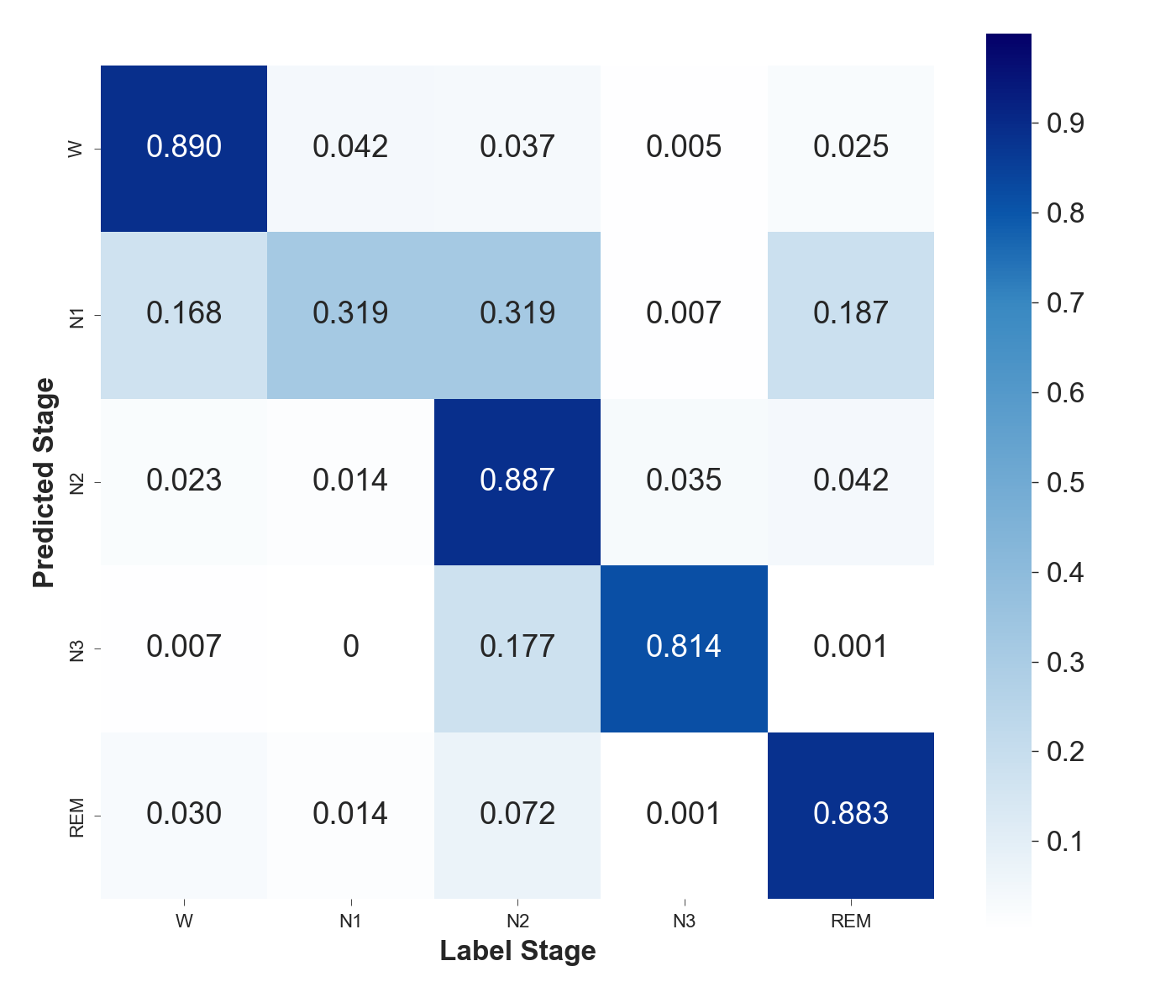}
      \caption{90\% Mask on Sleep-EDF-20.}
    \end{subfigure} \\

    \begin{subfigure}[b]{0.22\textwidth}
      \includegraphics[width=\linewidth]{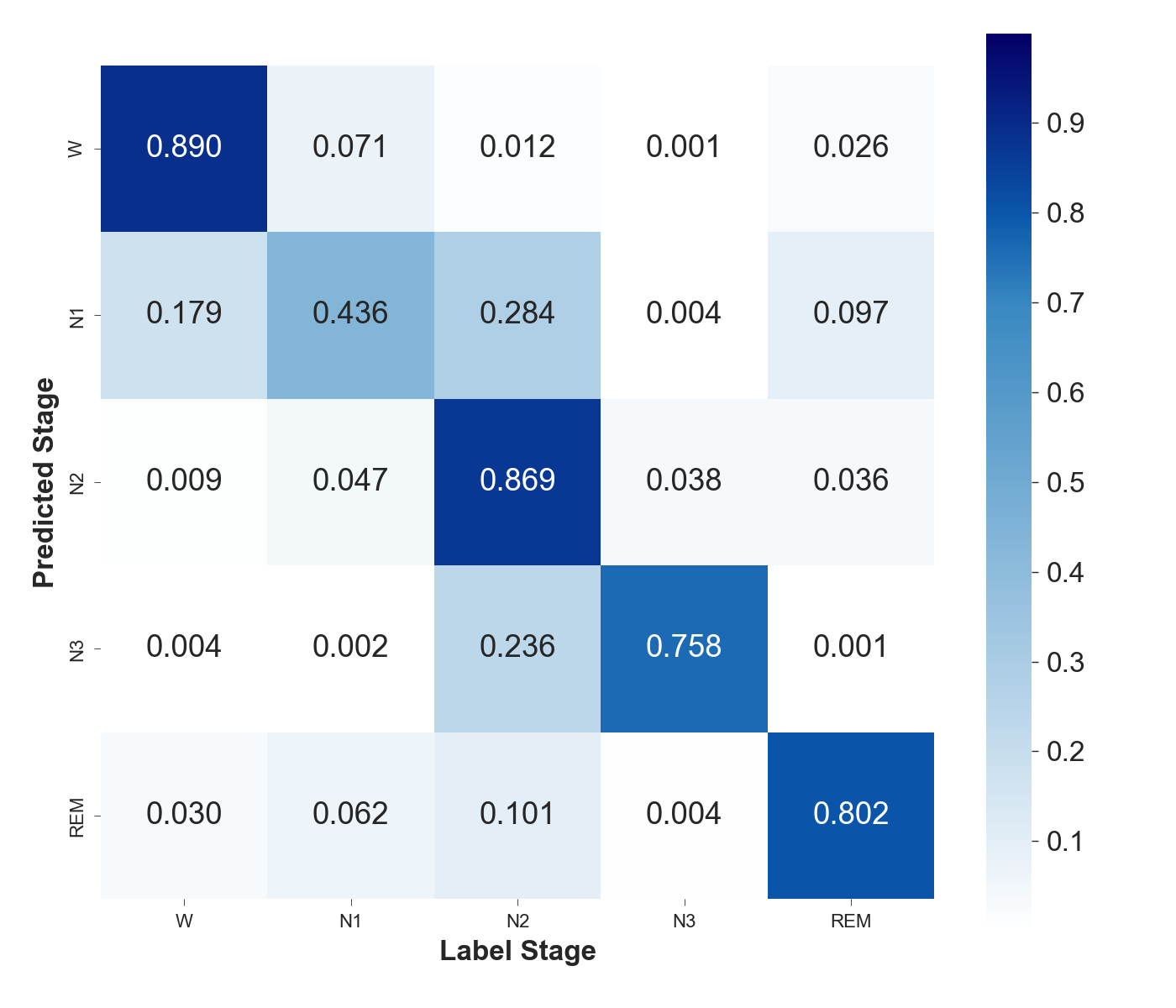}
      \caption{0\% Mask on Sleep-EDF-78.}
    \end{subfigure} &
    \begin{subfigure}[b]{0.22\textwidth}
      \includegraphics[width=\linewidth]{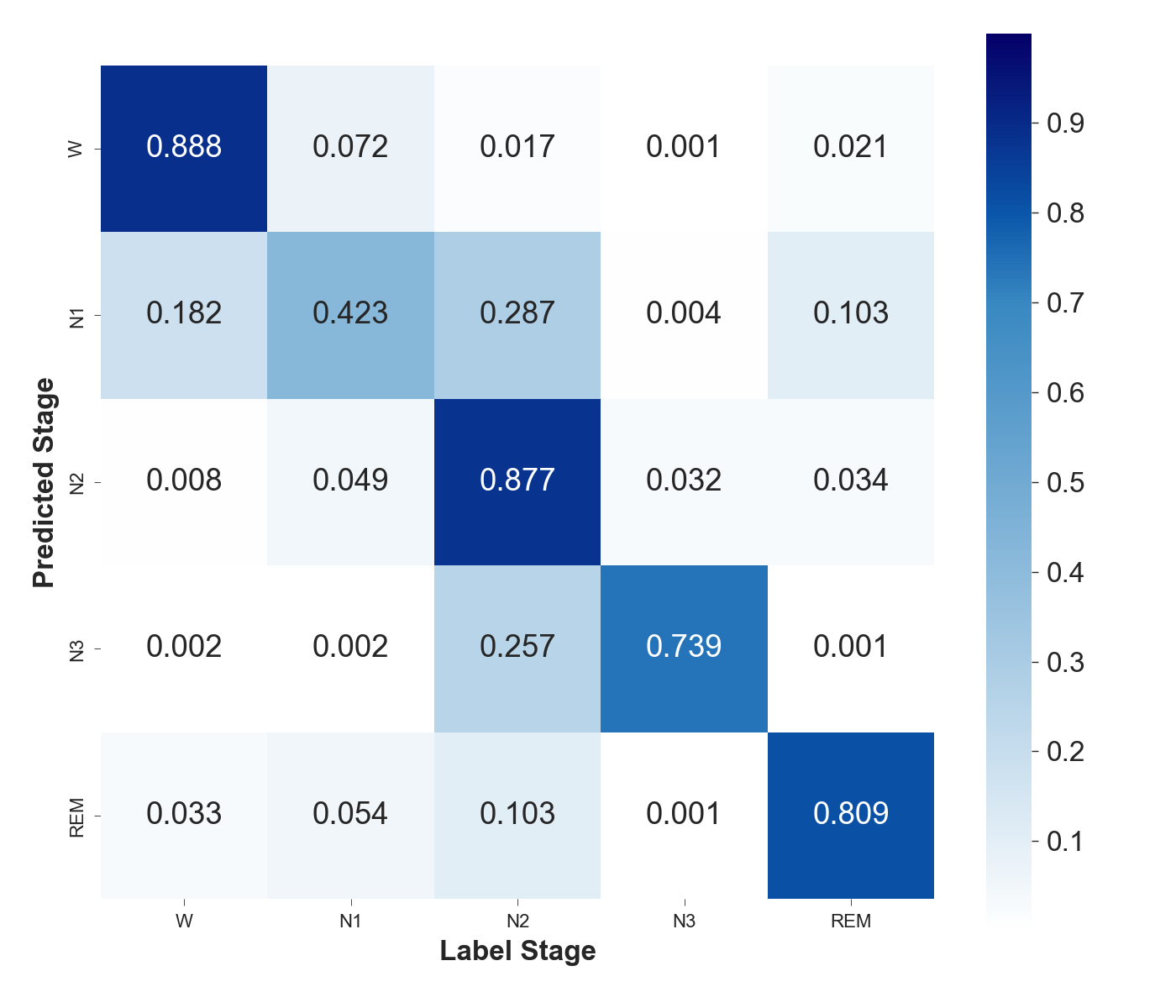}
      \caption{28\% Mask on Sleep-EDF-78.}
    \end{subfigure} &
    \begin{subfigure}[b]{0.22\textwidth}
      \includegraphics[width=\linewidth]{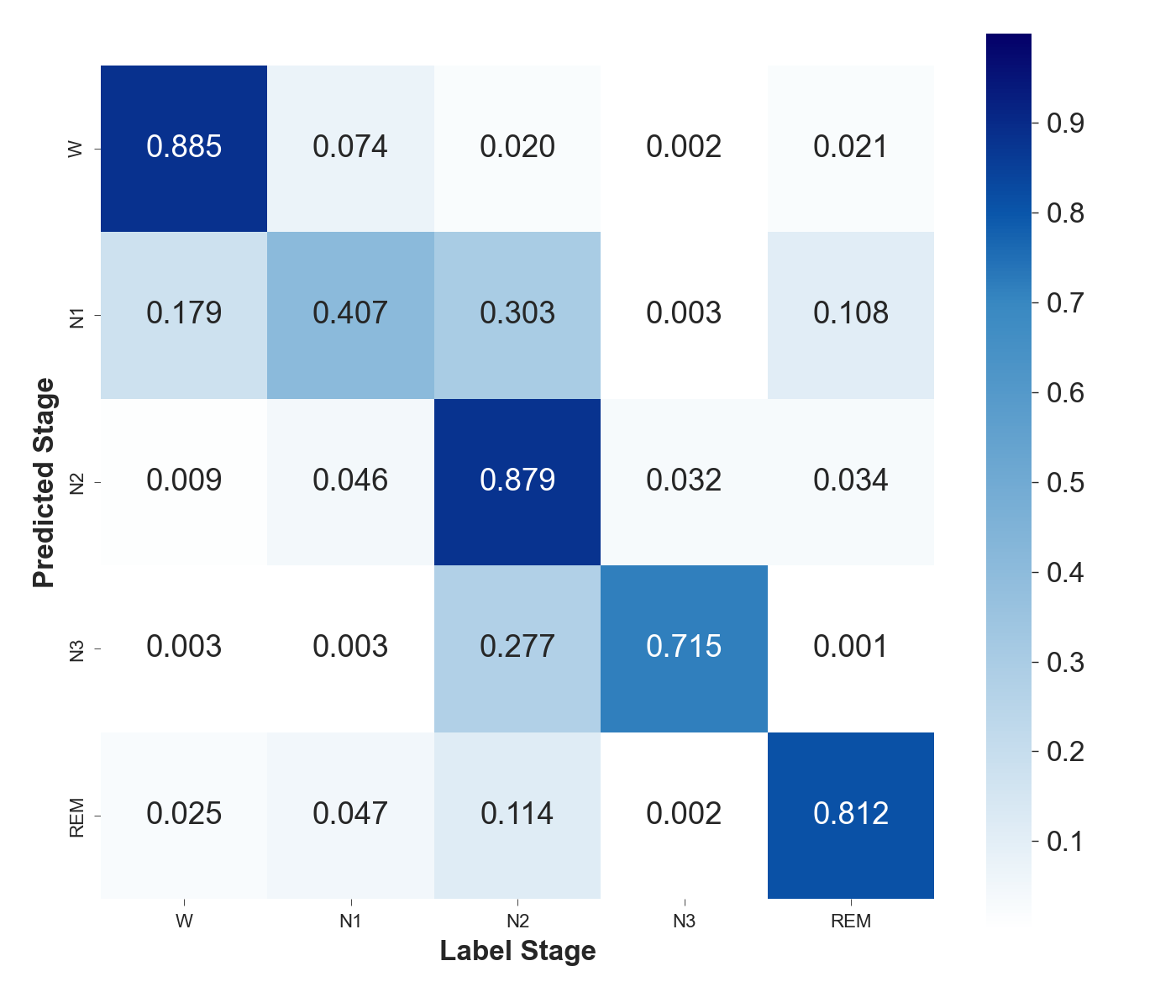}
      \caption{60\% Mask on Sleep-EDF-78.}
    \end{subfigure} &
    \begin{subfigure}[b]{0.22\textwidth}
      \includegraphics[width=\linewidth]{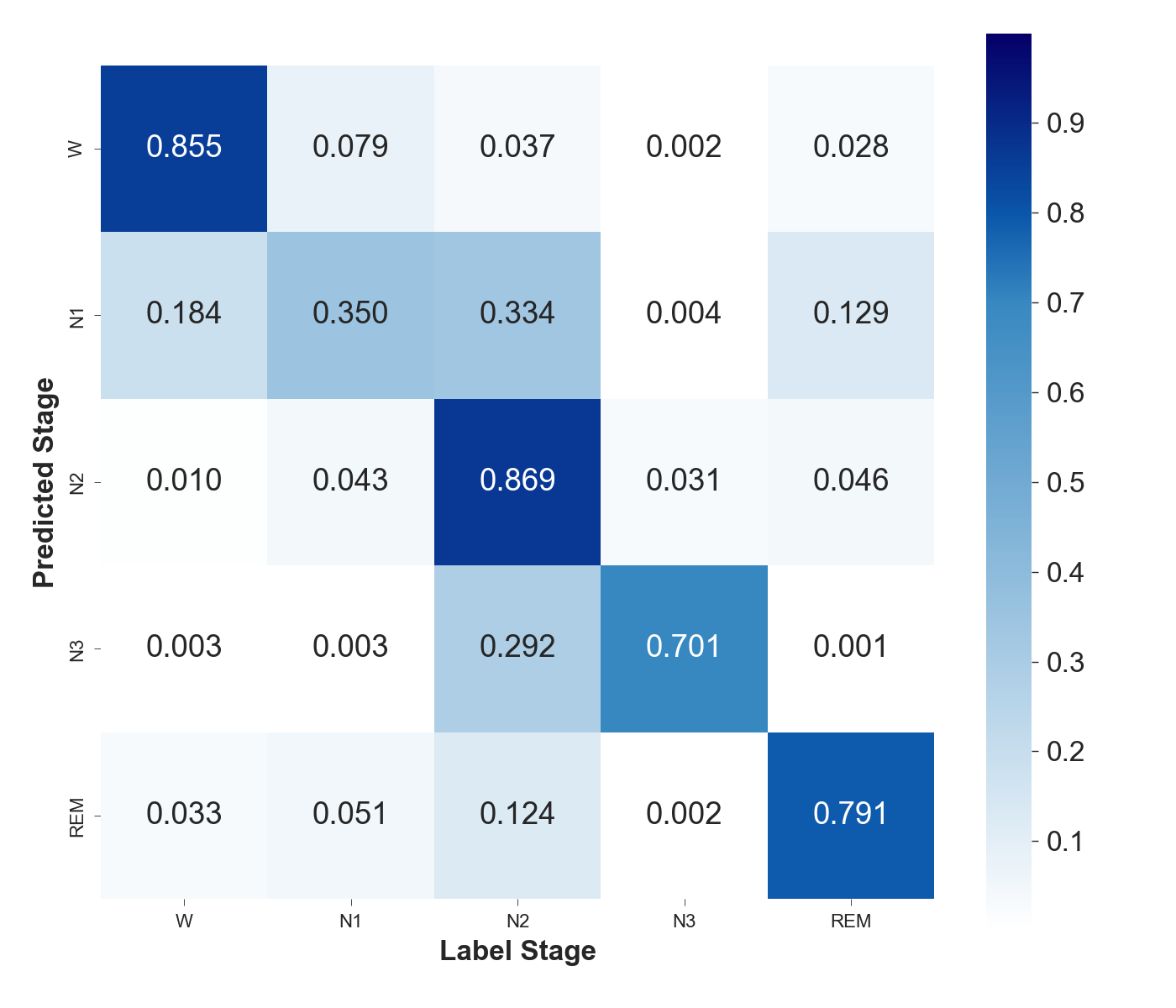}
      \caption{90\% Mask on Sleep-EDF-78.}
    \end{subfigure} \\

    \begin{subfigure}[b]{0.22\textwidth}
      \includegraphics[width=\linewidth]{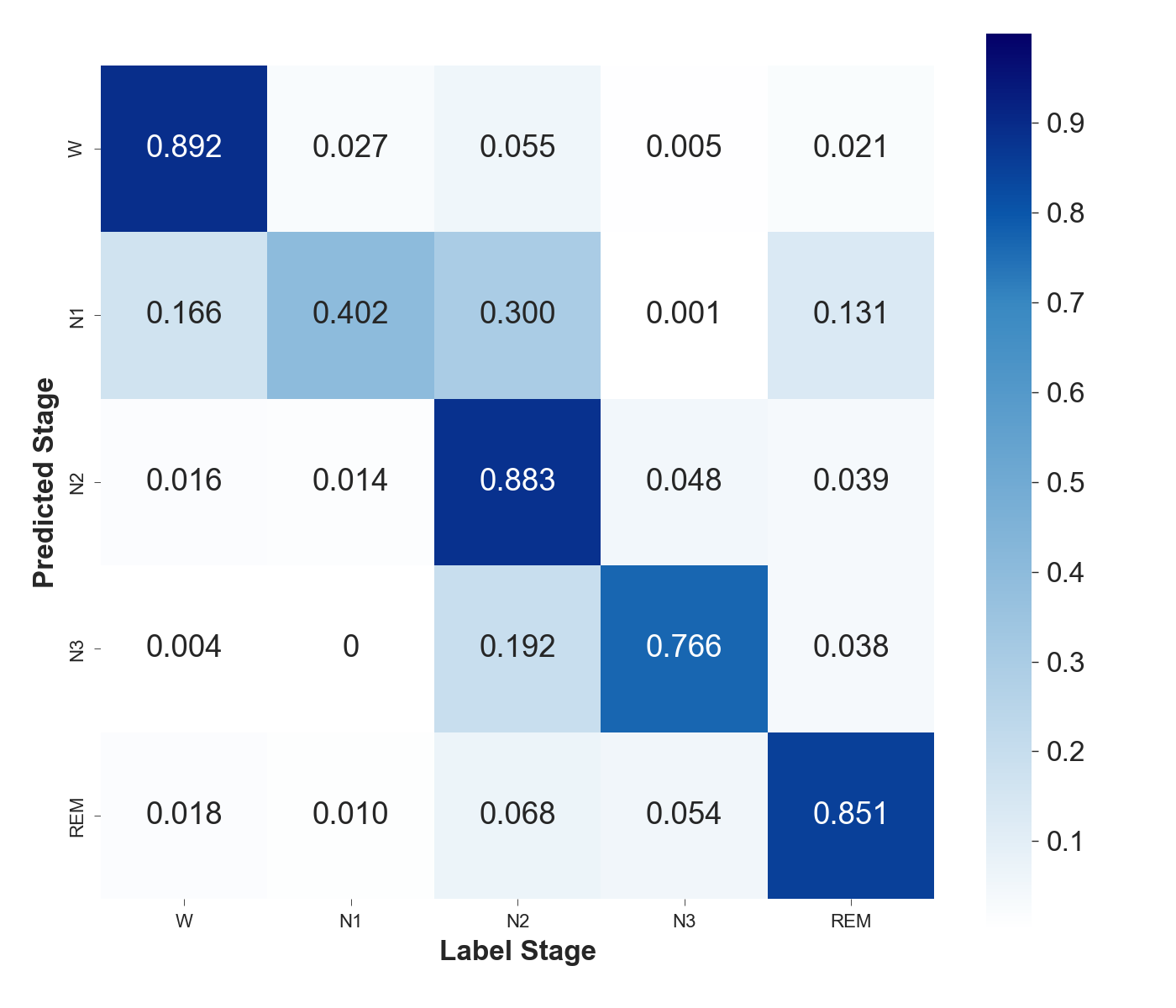}
      \caption{0\% Mask on SHHS.}
    \end{subfigure} &
    \begin{subfigure}[b]{0.22\textwidth}
      \includegraphics[width=\linewidth]{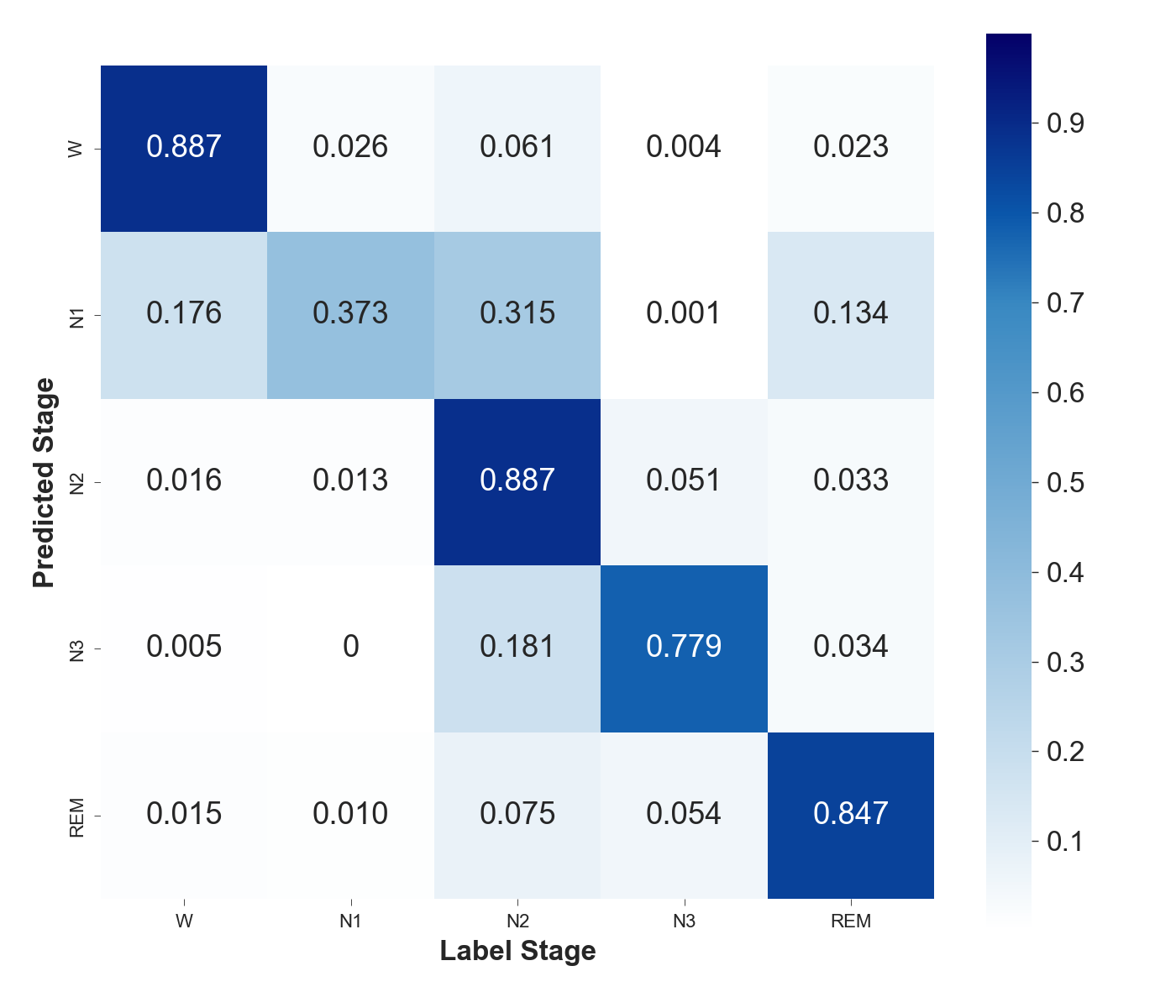}
      \caption{28\% Mask on SHHS.}
    \end{subfigure} &
    \begin{subfigure}[b]{0.22\textwidth}
      \includegraphics[width=\linewidth]{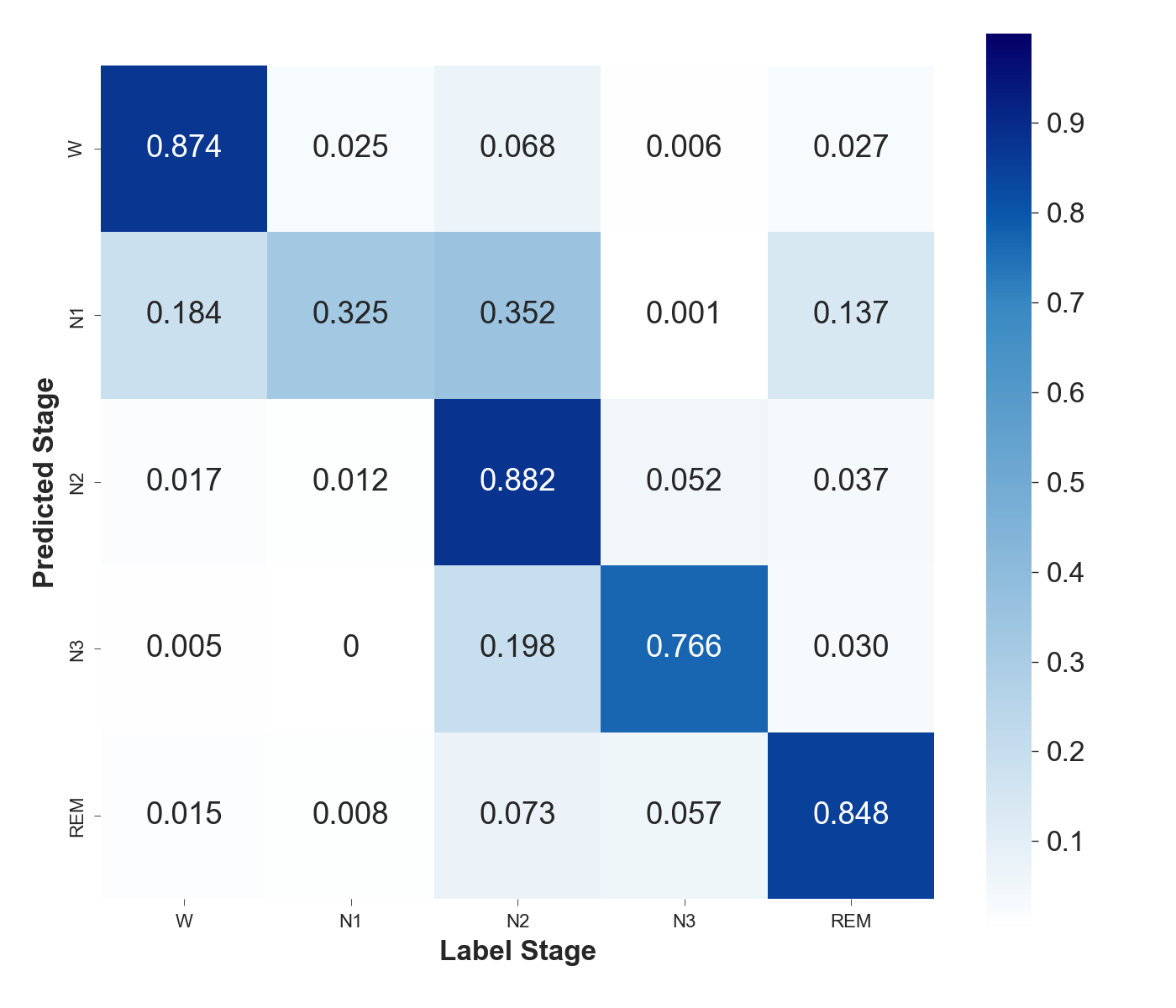}
      \caption{60\% Mask on SHHS.}
    \end{subfigure} &
    \begin{subfigure}[b]{0.22\textwidth}
      \includegraphics[width=\linewidth]{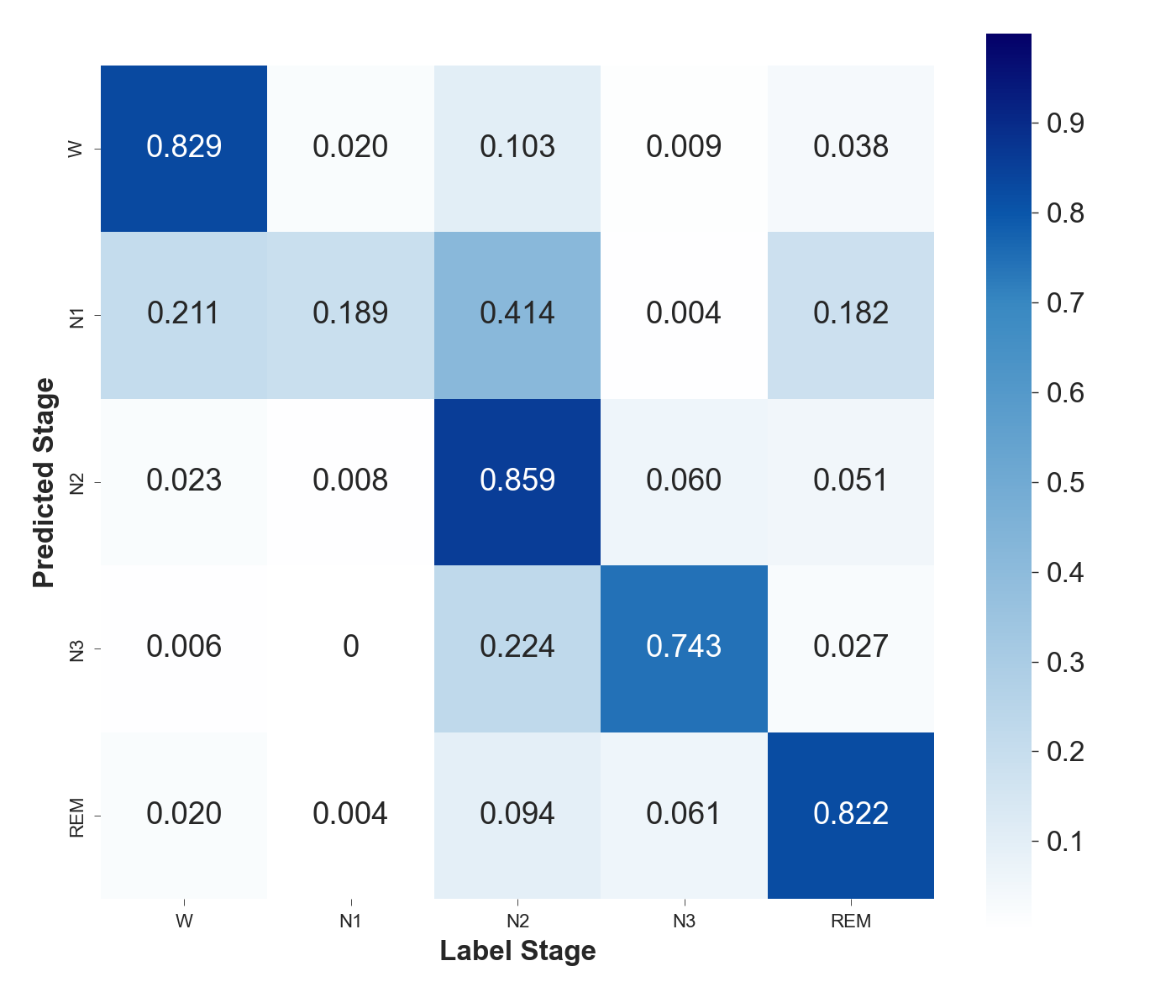}
      \caption{90\% Mask on SHHS.}
    \end{subfigure} \\
  \end{tabular}
  \caption{Confusion Matrix of Different Mask Ratios on Four Datasets.}
  \label{fig:4x4_grid}
\end{figure*}

For the sleep staging results of MASS under four different signal masking rates (under four different signal integrity levels), we have performed a confusion matrix visualization to analyze the results of MASS in more detail. The results are shown in Fig. \ref{fig:4x4_grid}. It is evident that among all sleep stages, the N1 stage consistently exhibits the lowest classification performance, with accuracy rarely exceeding 50\%. Furthermore, the N1 stage is particularly sensitive to signal loss, showing the steepest performance degradation as the masking ratio increases. For instance, under a 90\% masking rate, the classification accuracy of other stages typically decreases only within a few percentage points, whereas the N1 accuracy can experience a dramatic drop of several tens of percentage points. This observation aligns with findings from previous studies, which have noted that the N1 stage is inherently more difficult to classify due to its ambiguous EEG patterns and relatively limited sample size in most datasets, making it more vulnerable to performance fluctuations under partial observation scenarios. Despite this challenge, our proposed MASS framework demonstrates strong robustness, maintaining stable performance across all datasets even under extreme signal reduction. The relatively small overall decline in accuracy validates the effectiveness of our multi-level masking and prompt-based modeling design. However, the persistent underperformance on the N1 stage also reveals a critical limitation shared by existing methods, indicating a clear need for future research to develop more targeted strategies to better address this long-standing issue in automatic sleep staging.

\subsection{MASS on Different Mask Ratio}
To compare the performance of MASS under different mask ratios, we conducted more detailed mask ratio ablation experiments on four datasets. Specifically, we recorded 54 accuracy results of MASS at patch-level mask ratio $r_a$ from 0 to 0.8 and epoch-level mask ratio $r_e$ from 0 to 0.5 on each dataset, and the results are shown as follows. Across all four datasets, we observe a consistent trend: as the patch-level mask ratio $r_a$ and epoch-level mask ratio $r_e$ increase, model performance gradually declines. However, the decline is relatively smooth and mild, demonstrating the robustness of MASS to partial data availability. Notably, performance remains stable when $r_a \leq 0.5$ and $r_e \leq 0.2$, with accuracy drops typically within 1–2 percentage points compared to the full-signal setting. On the DREAMS-SUB dataset, MASS achieves the peak accuracy of 87.13\% at $r_a = 0.2$, $r_e = 0.0$, and maintains above 86\% accuracy even when $r_a = 0.5$, $r_e = 0.2$, indicating high resilience to moderate masking. On Sleep-EDF-20, a similar trend is observed. Accuracy peaks at 86.27\% under $r_a = 0.3$, $r_e = 0.0$, and remains above 85\% even with $r_a = 0.6$, $r_e = 0.2$, showing tolerance to more aggressive masking strategies. Sleep-EDF-78 demonstrates the flattest accuracy curve, with minimal fluctuations (within 1\%) across different mask ratios—suggesting the model’s robustness on larger-scale data. On SHHS, although the absolute performance drops more visibly under heavy masking (e.g., $r_a = 0.8, r_e = 0.5$), MASS still maintains accuracy above 80.5\%, highlighting its generalization to complex and diverse subjects. Overall, these results confirm that MASS not only performs well under full observation but also maintains high accuracy under a broad range of patch-level and epoch-level masking ratios. This robustness ensures its practical applicability in real-world wearable scenarios with unstable or limited data acquisition.

\begin{table*}[!htbp]
    \centering
    \adjustbox{width=\textwidth}
    {
    \begin{tabular}{cc|ccccccccc}
    \toprule
    \multirow{2}{*}{Accuracy}& &\multicolumn{9}{c}{$r_a$}\\
    &&0.0&0.1&0.2&0.3&0.4&0.5&0.6&0.7&0.8\\
    \midrule
    \multirow{6}{*}{$r_e$}&0.0&86.76&86.61&87.13&86.62&86.49&86.41&86.10&86.09&85.02\\
    &0.1&86.80&86.91&86.76&86.28&86.30&86.23&86.43&85.91&85.04\\
    &0.2&86.59&86.51&86.59&86.21&86.25&86.35&86.27&85.69&84.61\\    &0.3&86.28&86.19&86.41&85.90&85.86&85.71&85.44&84.86&84.42\\    &0.4&85.95&85.68&85.50&85.52&85.11&85.00&84.85&84.86&83.93\\    &0.5&85.18&85.13&85.04&84.78&85.11&84.72&84.26&84.13&83.31\\
    \bottomrule
    \end{tabular}
    }
    \caption{Ablation on different mask ratios on DREAMS-SUB dataset}
    \label{tab:mask_dreams}
\end{table*}

\begin{table*}[!htbp]
    \centering
    \adjustbox{width=\textwidth}
    {
    \begin{tabular}{cc|ccccccccc}
    \toprule
    \multirow{2}{*}{Accuracy}& &\multicolumn{9}{c}{$r_a$}\\
    &&0.0&0.1&0.2&0.3&0.4&0.5&0.6&0.7&0.8\\
    \midrule
    \multirow{6}{*}{$r_e$}&0.0&85.93&85.94&86.15&86.27&86.13&85.72&85.75&85.58&85.22\\
    &0.1&86.01&86.20&85.66&85.80&85.79&85.75&85.43&85.44&84.93\\
    &0.2&86.00&85.71&86.13&85.73&85.72&85.80&85.64&85.38&85.02\\    &0.3&85.68&85.83&85.56&85.70&85.59&85.58&85.34&84.83&84.59\\    &0.4&85.37&85.54&85.43&85.55&85.27&85.25&85.12&84.78&84.02\\    &0.5&85.20&85.19&84.93&85.05&84.87&84.98&84.60&84.32&83.80\\
    \bottomrule
    \end{tabular}
    }
    \caption{Ablation on different mask ratios on Sleep-EDF-20 dataset}
    \label{tab:mask_edf20}
\end{table*}

\begin{table*}[!htbp]
    \centering
    \adjustbox{width=\textwidth}
    {
    \begin{tabular}{cc|ccccccccc}
    \toprule
    \multirow{2}{*}{Accuracy}& &\multicolumn{9}{c}{$r_a$}\\
    &&0.0&0.1&0.2&0.3&0.4&0.5&0.6&0.7&0.8\\
    \midrule
    \multirow{6}{*}{$r_e$}&0.0&80.34&80.48&80.42&80.60&80.62&80.67&80.41&79.97&79.53\\
    &0.1&80.41&80.39&80.39&80.40&80.42&80.29&80.21&80.00&79.51\\
    &0.2&80.14&80.17&80.32&80.17&80.22&80.07&79.73&79.61&79.08\\    &0.3&79.94&80.05&80.10&80.11&79.88&79.95&79.69&79.40&78.80\\    &0.4&79.95&80.00&79.89&79.73&79.68&79.45&79.28&79.07&78.40\\    &0.5&79.56&79.41&79.46&79.39&79.13&79.08&78.94&78.55&77.90\\
    \bottomrule
    \end{tabular}
    }
    \caption{Ablation on different mask ratios on Sleep-EDF-78 dataset}
    \label{tab:mask_edf78}
\end{table*}

\begin{table*}[!htbp]
    \centering
    \adjustbox{width=\textwidth}
    {
    \begin{tabular}{cc|ccccccccc}
    \toprule
    \multirow{2}{*}{Accuracy}& &\multicolumn{9}{c}{$r_a$}\\
    &&0.0&0.1&0.2&0.3&0.4&0.5&0.6&0.7&0.8\\
    \midrule
    \multirow{6}{*}{$r_e$}&0.0&84.24&84.31&84.45&84.39&84.22&84.11&83.97&83.54&82.92\\
    &0.1&84.34&84.33&84.38&84.21&84.07&83.99&83.73&83.14&82.55\\
    &0.2&83.96&84.00&84.00&83.96&83.70&83.60&83.32&82.87&82.19\\    &0.3&83.79&83.74&83.77&83.56&83.32&83.19&83.03&82.36&81.67\\    &0.4&83.40&83.37&83.22&83.20&83.01&82.81&82.59&82.06&81.25\\    &0.5&82.90&82.77&82.77&82.73&82.49&82.31&81.94&81.39&80.57\\
    \bottomrule
    \end{tabular}
    }
    \caption{Ablation on different mask ratios on SHHS dataset}
    \label{tab:mask_shhs}
\end{table*}

\subsection{Detailed Comparison Experiments}
As shown in Table \ref{tab:Mask Comparison}, with complete inputs, the proposed MASS (Ours) model demonstrates competitive performance across all four datasets (DREAMS-SUB, Sleep-EDF-20, Sleep-EDF-78, and SHHS), achieving the highest Macro-F1 (MF1) score of 81.14\% on DREAMS-SUB and maintaining robust performance in other datasets. Notably, DeepSleepNet shows particularly strong results on Sleep-EDF-20 (MF1=81.36\%) and SHHS (MF1=80.37\%), while NeuroNet exhibits consistent performance across multiple datasets. \par
Table \ref{tab:Mask Comparison 2} shows the superiority of MASS with patch-level mask ratio and epoch-level mask ratio separatively equals to 0.2 and 0.1:
\begin{itemize}
    \item \textbf{Overall Superiority of MASS:} The results show that MASS consistently surpasses all other state-of-the-art models across all datasets. For example, in the DREAMS-SUB dataset, MASS achieves a per-class F1-score of 88.26\%, significantly higher than TinySleepNet's 73.32\%. This trend continues with scores of 88.90\%, 88.00\%, and 88.67\% in Sleep-EDF-20, Sleep-EDF-78, and SHHS, respectively.
    \item \textbf{Robustness Across Datasets:} MASS demonstrates strong and stable performance across various datasets. Unlike other methods that show score fluctuations, MASS consistently achieves high effectiveness, scoring above 80\% in overall accuracy across all datasets.
    \item \textbf{Superior Metrics Across the Board:} MASS excels in other metrics like macro F1 (MF1), Cohen's kappa ($\kappa$), and macro geometric mean (MGm). It records an MF1 of 86.75\% in DREAMS-SUB and 85.66\% in Sleep-EDF-20, the highest in those datasets. The kappa value of 0.81 in DREAMS-SUB indicates a strong agreement between predicted and actual classifications.
    \item \textbf{Comparison with other Models:} Traditional models, such as DeepSleepNet and CNN-Transformer-LSTM, consistently perform worse. For example, DeepSleepNet's highest F1-score is only 42.90\% in the SHHS dataset, underscoring the effectiveness of MASS compared to conventional methods.
\end{itemize} \par
Table \ref{tab:Mask Comparison 3} furtherly proves the advance of MASS in incomplete inputs, where patch-level mask ratio=0.5 and epoch-level mask ratio=0.2:
\begin{itemize}
    \item \textbf{Overall Superiority of MASS:} The results show that MASS significantly surpasses all other models across all datasets. For example, in the DREAMS-SUB dataset, MASS achieves a per-class F1-score of 89.89\%, much higher than TinySleepNet's best score of 66.21\%. This trend continues with scores of 89.52\%, 87.97\%, and 87.65\% in Sleep-EDF-20, Sleep-EDF-78, and SHHS, respectively.
    \item \textbf{Robustness Across Datasets:} MASS demonstrates strong and consistent results, achieving high scores above 80\% in overall accuracy (86.34\% in DREAMS-SUB, 85.80\% in Sleep-EDF-20). Other models show more variability in their performance.
    \item \textbf{Superior Metrics Across the Board:} MASS excels in overall metrics like MF1 and $\kappa$. It scores 86.34\% in MF1 for DREAMS-SUB and maintains high values in other datasets, indicating reliable classification. The kappa value of 0.80 reflects a strong agreement between predicted and actual classifications.
    \item \textbf{Comparison with other Models:} Other models, such as DeepSleepNet and LGSleepNet, consistently perform worse. For instance, DeepSleepNet's highest F1-score is only 36.07\% in the SHHS dataset, highlighting MASS's effectiveness.
\end{itemize} \par
Table \ref{tab:Mask Comparison 4} illustrates the performance of models in the condition of extremely incomplete inputs, where patch-level mask ratio=0.8 and epoch-level mask ratio=0.5. 
\begin{itemize}
    \item \textbf{Overall Superiority of MASS:} The results show that MASS is better than all other methods across all datasets. In the DREAMS-SUB dataset, MASS has a per-class F1-score of 84.93\%, while the next best, TinySleepNet, only scores 35.88\%. This pattern continues in other datasets, with scores of 87.69\%, 86.04\%, and 83.82\% for Sleep-EDF-20, Sleep-EDF-78, and SHHS.

    \item \textbf{Robustness Across Datasets:} MASS shows strong and consistent results across different datasets, achieving an accuracy (ACC) of 83.30\% in DREAMS-SUB and 83.80\% in Sleep-EDF-20, proving its reliability.

    \item \textbf{Superior Metrics Across the Board:} MASS also performs well in overall metrics like MF1 and $\kappa$. It has an MF1 of 75.57\% in DREAMS-SUB and maintains good scores in other datasets. The kappa value of 0.76 indicates a strong agreement between predicted and actual classifications.

    \item \textbf{Comparison with other Models:} Other models such as DeepSleepNet and LGSleepNet consistently perform worse. For example, DeepSleepNet's highest F1-score is just 22.74\% in the SHHS dataset, showcasing MASS's superior effectiveness.
\end{itemize}
In conclusion, the MASS method stands out as a highly effective tool for sleep staging, consistently outpacing existing state-of-the-art models across multiple datasets in resource-limited environment. Its high F1-scores, combined with robust overall metrics, underscore its potential for practical applications in sleep medicine. The results highlight that MASS not only enhances classification accuracy but also provides a reliable tool for researchers and clinicians. Future work should investigate the underlying mechanisms of MASS to further refine its effectiveness and expand its applicability in sleep analysis.

\begin{table*}[htbp]
\centering
\begin{adjustbox}{max width=0.9\linewidth}
\begin{tabular}{cc|ccccc|cccc} 
\toprule 
\multirow{2}{*}{Datasets}&\multirow{2}{*}{Method}&\multicolumn{5}{c|}{Per-Class F1-score}&\multicolumn{4}{c}{Overall Metrics} \\
\multirow{2}{*}{}&\multirow{2}{*}{}&W(\%)&N1(\%)&N2(\%)&N3(\%)&REM(\%) &ACC(\%)&MF1(\%)& $\kappa$ &MGm(\%) \\
\midrule 
\multirow{8}{*}{DREAMS-SUB}&DeepSleepNet&69.94&38.95&73.64&83.15&69.41&76.11&67.02&0.67&76.40 \\
\multirow{8}{*}{}&TinySleepNet&87.38&50.60&86.51&84.53&83.39&84.82&78.48&0.78&85.22 \\
\multirow{8}{*}{}&AttnSleep&86.11&38.13&85.60&\textbf{85.77}&76.74&82.39&74.47&0.75&82.86 \\
\multirow{8}{*}{}&LGSleepNet&87.83&36.78&85.76&83.70&77.47&82.46&74.31&0.75&82.62 \\
\multirow{8}{*}{}&CNN-Transformer-LSTM&62.42&0.00&77.84&79.23&56.63&71.01&55.22&0.59&62.69 \\
\multirow{8}{*}{}&NeuroNet&88.90&50.13&\textbf{87.98}&84.90&85.65&86.36&79.51&0.80&85.69 \\
\multirow{8}{*}{}&\textbf{MASS(Ours)}&\textbf{89.17}&\textbf{57.38}&87.62&83.97&\textbf{87.56}&\textbf{86.76}&\textbf{81.14}&\textbf{0.81}&\textbf{87.29} \\
\midrule 
\multirow{8}{*}{Sleep-EDF-20}&DeepSleepNet&87.05&\textbf{55.94}&\textbf{89.56}&87.81&86.47&\textbf{86.75}&\textbf{81.36}&\textbf{0.81}&\textbf{88.03} \\
\multirow{8}{*}{}&TinySleepNet&86.90&47.32&87.93&86.74&83.15&84.70&78.41&0.78&86.01 \\
\multirow{8}{*}{}&AttnSleep&87.97&43.57&88.67&88.84&80.50&84.21&77.91&0.77&86.20 \\
\multirow{8}{*}{}&LGSleepNet&87.59&41.87&88.43&\textbf{89.24}&79.67&83.90&77.36&0.77&85.59 \\
\multirow{8}{*}{}&CNN-Transformer-LSTM&80.45&0.00&82.95&85.20&64.15&76.66&62.55&0.67&68.80 \\
\multirow{8}{*}{}&NeuroNet&88.19&44.31&88.51&88.10&84.15&85.80&78.65&0.79&85.34 \\
\multirow{8}{*}{}&\textbf{MASS(Ours)}&\textbf{89.99}&51.19&87.93&84.28&\textbf{87.11}&85.93&80.10&0.80&87.38 \\
\midrule 
\multirow{8}{*}{Sleep-EDF-78}&DeepSleepNet&\textbf{90.02}&\textbf{54.28}&\textbf{87.21}&\textbf{81.12}&\textbf{86.13}&\textbf{83.75}&\textbf{79.75}&\textbf{0.77}&\textbf{87.08} \\
\multirow{8}{*}{}&TinySleepNet&87.43&46.40&85.41&78.60&75.68&79.57&74.71&0.71&83.49 \\
\multirow{8}{*}{}&AttnSleep&87.01&42.99&84.67&80.00&74.80&78.47&73.90&0.70&82.76 \\
\multirow{8}{*}{}&LGSleepNet&86.39&43.26&84.53&79.45&74.23&78.03&73.57&0.69&82.81 \\
\multirow{8}{*}{}&CNN-Transformer-LSTM&86.46&34.30&82.38&74.36&65.81&75.89&68.66&0.66&78.88 \\
\multirow{8}{*}{}&NeuroNet&88.05&46.99&85.82&77.92&79.83&81.05&75.72&0.73&83.61 \\
\multirow{8}{*}{}&\textbf{MASS(Ours)}&88.16&48.73&84.96&73.65&79.55&80.33&75.01&0.72&83.51 \\
\midrule 
\multirow{8}{*}{SHHS}&DeepSleepNet&\textbf{90.70}&\textbf{52.69}&\textbf{89.21}&81.05&\textbf{88.17}&\textbf{86.73}&\textbf{80.37}&\textbf{0.81}&\textbf{87.20} \\
\multirow{8}{*}{}&TinySleepNet&86.17&40.69&86.76&81.20&81.25&83.29&75.21&0.76&83.07 \\
\multirow{8}{*}{}&AttnSleep&85.62&33.07&86.90&79.81&79.26&81.72&72.93&0.74&82.86 \\
\multirow{8}{*}{}&LGSleepNet&86.70&30.90&87.13&\textbf{81.65}&79.62&82.53&73.20&0.75&82.60 \\
\multirow{8}{*}{}&CNN-Transformer-LSTM&84.03&34.99&85.33&80.08&79.32&81.71&72.75&0.74&80.92 \\
\multirow{8}{*}{}&NeuroNet&87.84&43.03&88.27&80.80&84.40&84.94&76.87&0.78&83.91 \\
\multirow{8}{*}{}&\textbf{MASS(Ours)}&88.91&45.17&86.96&77.91&85.39&84.24&76.87&0.77&84.20 \\
\bottomrule 
\end{tabular}
\end{adjustbox}
\caption{Comparison with State-of-The-Art models on DREAMS-SUB, Sleep-EDF-20, Sleep-EDF-78 and SHHS datasets with 
\textbf{Patch-Level Mask Ratio=0.0} and \textbf{Epoch-Level Mask Ratio=0.0}.} 
\label{tab:Mask Comparison}
\end{table*}

\begin{table*}[htbp]
\centering
\begin{adjustbox}{max width=0.9\linewidth}
\begin{tabular}{cc|ccccc|cccc} 
\toprule 
\multirow{2}{*}{Datasets}&\multirow{2}{*}{Method}&\multicolumn{5}{c|}{Per-Class F1-score}&\multicolumn{4}{c}{Overall Metrics} \\
\multirow{2}{*}{}&\multirow{2}{*}{}&W(\%)&N1(\%)&N2(\%)&N3(\%)&REM(\%) &ACC(\%)&MF1(\%)& $\kappa$ &MGm(\%) \\
\midrule 
\multirow{8}{*}{DREAMS-SUB}&DeepSleepNet&14.04&5.37&25.65&14.81&12.21&22.31&14.42&0.01&27.18 \\
\multirow{8}{*}{}&TinySleepNet&73.32&32.66&74.46&65.89&66.04&69.77&62.47&0.58&75.23 \\
\multirow{8}{*}{}&AttnSleep&78.28&27.30&75.84&55.07&60.27&69.02&59.35&0.56&71.37 \\
\multirow{8}{*}{}&LGSleepNet&79.69&26.15&73.17&48.55&68.42&67.57&59.19&0.54&71.38 \\
\multirow{8}{*}{}&CNN-Transformer-LSTM&53.38&0.09&70.40&63.81&50.15&61.53&47.56&0.46&57.32 \\
\multirow{8}{*}{}&NeuroNet&67.14&32.39&72.39&55.58&45.63&64.76&54.63&0.50&67.92 \\
\multirow{8}{*}{}&\textbf{MASS(Ours)}&\textbf{88.26}&\textbf{55.32}&\textbf{87.71}&\textbf{83.73}&\textbf{88.03}&\textbf{86.75}&\textbf{80.61}&\textbf{0.81}&\textbf{86.61} \\
\midrule 
\multirow{8}{*}{Sleep-EDF-20}&DeepSleepNet&11.33&6.99&25.64&11.53&13.78&20.28&13.86&0.00&29.40 \\
\multirow{8}{*}{}&TinySleepNet&82.56&28.90&78.87&72.01&63.48&70.47&65.16&0.60&77.89 \\
\multirow{8}{*}{}&AttnSleep&81.68&34.19&80.71&50.85&69.49&72.62&63.38&0.61&74.85 \\
\multirow{8}{*}{}&LGSleepNet&76.35&29.87&77.24&44.50&65.29&69.83&58.65&0.57&68.42 \\
\multirow{8}{*}{}&CNN-Transformer-LSTM&74.05&0.00&75.92&67.36&57.36&68.43&54.94&0.55&61.76 \\
\multirow{8}{*}{}&NeuroNet&79.07&29.34&76.23&43.83&56.75&66.51&57.04&0.52&69.46 \\
\multirow{8}{*}{}&\textbf{MASS(Ours)}&\textbf{88.90}&\textbf{48.37}&\textbf{87.99}&\textbf{83.83}&\textbf{87.16}&\textbf{85.66}&\textbf{79.25}&\textbf{0.79}&\textbf{86.15} \\
\midrule 
\multirow{8}{*}{Sleep-EDF-78}&DeepSleepNet&27.43&19.47&36.85&17.28&16.18&31.07&23.44&0.10&38.74 \\
\multirow{8}{*}{}&TinySleepNet&82.92&43.71&81.29&66.26&67.02&73.17&68.24&0.63&78.33 \\
\multirow{8}{*}{}&AttnSleep&82.83&38.90&79.69&39.60&63.00&70.67&60.80&0.58&71.34 \\
\multirow{8}{*}{}&LGSleepNet&74.98&27.03&63.09&24.29&57.58&55.72&49.39&0.41&63.24 \\
\multirow{8}{*}{}&CNN-Transformer-LSTM&79.26&32.30&70.32&59.64&59.29&65.46&60.16&0.53&73.22 \\
\multirow{8}{*}{}&NeuroNet&81.56&35.11&76.39&43.31&64.75&69.44&60.22&0.57&71.23 \\
\multirow{8}{*}{}&\textbf{MASS(Ours)}&\textbf{88.00}&\textbf{47.66}&\textbf{85.05}&\textbf{73.83}&\textbf{80.20}&\textbf{80.38}&\textbf{74.95}&\textbf{0.72}&\textbf{83.18} \\
\midrule 
\multirow{8}{*}{SHHS}&DeepSleepNet&42.90&10.04&37.26&31.37&28.31&33.48&29.98&0.17&46.05 \\
\multirow{8}{*}{}&TinySleepNet&80.96&34.10&79.83&61.60&70.62&74.10&65.42&0.62&75.55 \\
\multirow{8}{*}{}&AttnSleep&79.57&15.72&77.48&43.73&68.45&70.24&56.99&0.56&66.66 \\
\multirow{8}{*}{}&LGSleepNet&76.72&11.57&78.41&52.92&66.55&71.04&57.23&0.56&65.74 \\
\multirow{8}{*}{}&CNN-Transformer-LSTM&65.73&26.20&75.00&49.05&73.51&68.35&57.90&0.55&70.41 \\
\multirow{8}{*}{}&NeuroNet&72.20&24.09&76.79&49.44&67.17&69.59&57.93&0.55&68.41 \\
\multirow{8}{*}{}&\textbf{MASS(Ours)}&\textbf{88.67}&\textbf{43.01}&\textbf{87.09}&\textbf{78.37}&\textbf{85.74}&\textbf{84.37}&\textbf{76.58}&\textbf{0.77}&\textbf{83.85} \\
\bottomrule 
\end{tabular}
\end{adjustbox}
\caption{Comparison with State-of-The-Art models on DREAMS-SUB, Sleep-EDF-20, Sleep-EDF-78 and SHHS datasets with \textbf{Patch-Level Mask Ratio=0.2} and \textbf{Epoch-Level Mask Ratio=0.1}.} 
\label{tab:Mask Comparison 2}
\end{table*}

\begin{table*}[htbp]
\centering
\begin{adjustbox}{max width=0.9\linewidth}
\begin{tabular}{cc|ccccc|cccc} 
\toprule 
\multirow{2}{*}{Datasets}&\multirow{2}{*}{Method}&\multicolumn{5}{c|}{Per-Class F1-score}&\multicolumn{4}{c}{Overall Metrics} \\
\multirow{2}{*}{}&\multirow{2}{*}{}&W(\%)&N1(\%)&N2(\%)&N3(\%)&REM(\%) &ACC(\%)&MF1(\%)& $\kappa$ &MGm(\%) \\
\midrule 
\multirow{8}{*}{DREAMS-SUB}&DeepSleepNet&14.07&5.17&25.45&12.17&9.97&21.80&13.37&0.00&24.62 \\
\multirow{8}{*}{}&TinySleepNet&66.21&21.41&58.37&25.23&41.97&51.56&42.64&0.35&59.07 \\
\multirow{8}{*}{}&AttnSleep&44.33&10.60&53.21&2.22&27.25&40.29&27.52&0.22&40.15 \\
\multirow{8}{*}{}&LGSleepNet&49.41&12.48&49.18&3.91&41.77&41.39&31.35&0.24&48.12 \\
\multirow{8}{*}{}&CNN-Transformer-LSTM&40.14&0.28&47.60&24.79&41.73&43.29&30.91&0.25&41.78 \\
\multirow{8}{*}{}&NeuroNet&45.59&12.89&45.53&15.39&17.92&39.42&27.47&0.20&41.80 \\
\multirow{8}{*}{}&\textbf{MASS(Ours)}&\textbf{89.89}&\textbf{51.47}&\textbf{87.26}&\textbf{83.07}&\textbf{88.35}&\textbf{86.34}&\textbf{80.01}&\textbf{0.80}&\textbf{85.91} \\
\midrule 
\multirow{8}{*}{Sleep-EDF-20}&DeepSleepNet&8.42&5.80&24.33&11.94&13.32&20.17&12.76&0.00&26.82 \\
\multirow{8}{*}{}&TinySleepNet&71.73&22.74&69.05&27.92&29.15&52.53&44.12&0.37&60.06 \\
\multirow{8}{*}{}&AttnSleep&73.37&27.99&72.11&2.48&44.78&56.93&44.15&0.41&58.88 \\
\multirow{8}{*}{}&LGSleepNet&69.10&23.70&70.43&3.94&48.41&57.37&43.11&0.40&55.08 \\
\multirow{8}{*}{}&CNN-Transformer-LSTM&46.71&0.00&46.04&10.21&43.68&43.49&29.33&0.24&37.89 \\
\multirow{8}{*}{}&NeuroNet&58.05&22.44&70.31&4.26&13.43&49.04&33.70&0.29&47.88 \\
\multirow{8}{*}{}&\textbf{MASS(Ours)}&\textbf{89.52}&\textbf{47.77}&\textbf{88.04}&\textbf{83.61}&\textbf{87.40}&\textbf{85.80}&\textbf{79.27}&\textbf{0.80}&\textbf{86.24} \\
\midrule 
\multirow{8}{*}{Sleep-EDF-78}&DeepSleepNet&24.32&16.18&36.29&8.72&12.42&29.07&19.59&0.06&33.66 \\
\multirow{8}{*}{}&TinySleepNet&74.42&38.87&75.66&18.20&55.29&63.32&52.49&0.49&65.16 \\
\multirow{8}{*}{}&AttnSleep&73.37&28.55&74.09&1.34&48.36&61.87&45.14&0.45&55.75 \\
\multirow{8}{*}{}&LGSleepNet&65.36&16.07&58.01&3.41&46.59&49.48&37.89&0.31&50.72 \\
\multirow{8}{*}{}&CNN-Transformer-LSTM&68.68&23.85&25.55&7.34&35.21&40.94&32.13&0.26&48.78 \\
\multirow{8}{*}{}&NeuroNet&69.71&20.00&66.12&3.49&47.68&57.07&41.40&0.40&53.67 \\
\multirow{8}{*}{}&\textbf{MASS(Ours)}&\textbf{87.97}&\textbf{46.47}&\textbf{84.59}&\textbf{71.79}&\textbf{80.27}&\textbf{80.07}&\textbf{74.22}&\textbf{0.72}&\textbf{82.41} \\
\midrule 
\multirow{8}{*}{SHHS}&DeepSleepNet&36.07&7.41&30.21&15.67&22.84&27.19&22.44&0.08&38.21 \\
\multirow{8}{*}{}&TinySleepNet&76.07&22.92&70.11&11.42&55.68&61.47&47.24&0.41&58.10 \\
\multirow{8}{*}{}&AttnSleep&67.76&2.39&68.88&3.48&41.83&57.47&36.87&0.33&43.95 \\
\multirow{8}{*}{}&LGSleepNet&53.31&2.22&68.04&25.78&40.24&56.03&37.92&0.33&46.86 \\
\multirow{8}{*}{}&CNN-Transformer-LSTM&51.01&9.49&54.69&5.40&51.78&47.71&34.47&0.29&49.38 \\
\multirow{8}{*}{}&NeuroNet&50.36&3.71&63.50&24.49&25.76&49.77&33.57&0.27&46.19 \\
\multirow{8}{*}{}&\textbf{MASS(Ours)}&\textbf{87.65}&\textbf{39.25}&\textbf{86.31}&\textbf{77.20}&\textbf{85.48}&\textbf{83.59}&\textbf{75.18}&\textbf{0.76}&\textbf{82.58} \\
\bottomrule 
\end{tabular}
\end{adjustbox}
\caption{Comparison with State-of-The-Art models on DREAMS-SUB, Sleep-EDF-20, Sleep-EDF-78 and SHHS datasets with \textbf{Patch-Level Mask Ratio=0.5} and \textbf{Epoch-Level Mask Ratio=0.2}.} 
\label{tab:Mask Comparison 3}
\end{table*}

\begin{table*}[htbp]
\centering
\begin{adjustbox}{max width=0.9\linewidth}
\begin{tabular}{cc|ccccc|cccc} 
\toprule 
\multirow{2}{*}{Datasets}&\multirow{2}{*}{Method}&\multicolumn{5}{c|}{Per-Class F1-score}&\multicolumn{4}{c}{Overall Metrics} \\
\multirow{2}{*}{}&\multirow{2}{*}{}&W(\%)&N1(\%)&N2(\%)&N3(\%)&REM(\%) &ACC(\%)&MF1(\%)& $\kappa$ &MGm(\%) \\
\midrule 
\multirow{8}{*}{DREAMS-SUB}&DeepSleepNet&9.71&3.96&27.06&6.36&8.26&23.50&11.07&0.00&16.82 \\
\multirow{8}{*}{}&TinySleepNet&35.88&6.19&1.73&0.00&0.22&19.29&8.80&0.03&16.74 \\
\multirow{8}{*}{}&AttnSleep&29.21&0.00&0.62&0.00&0.11&17.81&5.98&0.00&1.41 \\
\multirow{8}{*}{}&LGSleepNet&29.00&1.75&1.35&0.00&1.00&17.76&6.62&0.00&5.29 \\
\multirow{8}{*}{}&CNN-Transformer-LSTM&22.69&0.00&8.30&0.23&22.56&20.36&10.76&0.03&15.62 \\
\multirow{8}{*}{}&NeuroNet&30.25&0.08&2.49&0.18&0.42&18.90&6.68&0.00&3.74 \\
\multirow{8}{*}{}&\textbf{MASS(Ours)}&\textbf{84.93}&\textbf{42.38}&\textbf{84.42}&\textbf{80.88}&\textbf{85.24}&\textbf{83.30}&\textbf{75.57}&\textbf{0.76}&\textbf{82.40} \\
\midrule 
\multirow{8}{*}{Sleep-EDF-20}&DeepSleepNet&5.73&3.79&19.16&12.09&10.55&19.91&10.26&0.00&18.66 \\
\multirow{8}{*}{}&TinySleepNet&19.00&13.47&15.91&0.35&0.49&13.11&9.85&0.03&19.18 \\
\multirow{8}{*}{}&AttnSleep&42.93&14.93&22.23&0.00&1.56&18.61&16.33&0.08&27.66 \\
\multirow{8}{*}{}&LGSleepNet&45.82&16.85&43.74&0.00&12.47&31.82&23.77&0.15&38.52 \\
\multirow{8}{*}{}&CNN-Transformer-LSTM&4.94&0.00&4.35&0.00&30.53&20.48&7.96&0.00&5.26 \\
\multirow{8}{*}{}&NeuroNet&4.41&13.51&23.80&0.16&3.01&15.11&8.98&0.02&17.49 \\
\multirow{8}{*}{}&\textbf{MASS(Ours)}&\textbf{87.69}&\textbf{39.58}&\textbf{86.35}&\textbf{82.09}&\textbf{84.87}&\textbf{83.80}&\textbf{76.12}&\textbf{0.77}&\textbf{83.35} \\
\midrule 
\multirow{8}{*}{Sleep-EDF-78}&DeepSleepNet&16.32&10.45&29.79&4.13&9.14&26.21&13.97&0.01&23.54 \\
\multirow{8}{*}{}&TinySleepNet&31.61&27.05&55.47&0.05&20.14&39.38&26.86&0.20&39.62 \\
\multirow{8}{*}{}&AttnSleep&51.96&6.64&49.39&0.00&15.61&42.39&24.72&0.20&33.10 \\
\multirow{8}{*}{}&LGSleepNet&43.63&3.68&36.52&1.27&16.97&34.84&20.41&0.13&29.46 \\
\multirow{8}{*}{}&CNN-Transformer-LSTM&49.20&13.57&0.50&0.00&8.29&25.17&14.31&0.07&19.33 \\
\multirow{8}{*}{}&NeuroNet&46.52&3.79&37.04&0.01&13.50&35.83&20.17&0.13&29.34 \\
\multirow{8}{*}{}&\textbf{MASS(Ours)}&\textbf{86.04}&\textbf{40.79}&\textbf{83.04}&\textbf{71.31}&\textbf{76.85}&\textbf{77.89}&\textbf{71.61}&\textbf{0.69}&\textbf{80.48} \\
\midrule 
\multirow{8}{*}{SHHS}&DeepSleepNet&22.74&2.43&13.84&6.07&19.30&21.69&12.88&0.01&23.24 \\
\multirow{8}{*}{}&TinySleepNet&44.42&0.80&50.40&0.00&36.79&41.11&26.48&0.18&36.76 \\
\multirow{8}{*}{}&AttnSleep&27.59&0.02&22.46&0.22&1.33&22.29&10.33&0.04&12.27 \\
\multirow{8}{*}{}&LGSleepNet&31.53&0.29&22.53&15.06&5.89&26.92&15.06&0.07&19.95 \\
\multirow{8}{*}{}&CNN-Transformer-LSTM&27.96&0.16&5.27&376.50&6.05&17.42&7.89&0.01&9.50 \\
\multirow{8}{*}{}&NeuroNet&27.57&0.07&13.27&9.32&3.02&19.95&10.65&0.03&16.44 \\
\multirow{8}{*}{}&\textbf{MASS(Ours)}&\textbf{83.82}&\textbf{26.87}&\textbf{83.51}&\textbf{74.71}&\textbf{82.29}&\textbf{80.57}&\textbf{70.24}&\textbf{0.72}&\textbf{78.21} \\
\bottomrule 
\end{tabular}
\end{adjustbox}
\caption{Comparison with State-of-The-Art models on DREAMS-SUB, Sleep-EDF-20, Sleep-EDF-78 and SHHS datasets with \textbf{Patch-Level Mask Ratio=0.8} and \textbf{Epoch-Level Mask Ratio=0.5}.} 
\label{tab:Mask Comparison 4}
\end{table*}

\end{document}